\documentclass[11pt]{article}

\usepackage{ACL2023}

\usepackage{times}
\usepackage{latexsym}

\usepackage[T1]{fontenc}
\usepackage[utf8]{inputenc}

\usepackage{microtype}

\usepackage{inconsolata}
\usepackage{booktabs}
\usepackage{graphicx}

\usepackage{multirow}

\usepackage{colortbl} 
\usepackage{subfigure}
\usepackage{subcaption}

\usepackage{graphicx}

\usepackage{amsmath}

\usepackage{xcolor}
\usepackage{amssymb}
\usepackage{pifont}
\newcommand{\cmark}{\ding{51}}%
\newcommand{\xmark}{\ding{55}}%
\usepackage[ruled]{algorithm2e}

\usepackage{tikz}
\newcommand*\circled[1]{\tikz[baseline=(char.base)]{
            \node[shape=circle,draw,inner sep=0.7pt] (char) {#1};}}

\title{GeoEval: Benchmark for Evaluating LLMs and Multi-Modal Models on Geometry Problem-Solving}

\author{
    Jiaxin Zhang\textsuperscript{\(\S\)}\thanks{~~~Equal Contribution} , Zhong-Zhi Li\textsuperscript{\(\diamond\)}\footnotemark[1] , Ming-Liang Zhang\textsuperscript{\(\diamond\)}, Fei Yin\textsuperscript{\(\circledast \)\(\diamond\)}, Cheng-Lin Liu\textsuperscript{\(\circledast\)\(\diamond\)}\footnotemark[2] , Yashar Moshfeghi\textsuperscript{\(\S\)}\thanks{~~~Corresponding Author} \\
    School of Artificial Intelligence, University of Chinese Academy of Sciences\textsuperscript{\(\diamond\)} \\
    MAIS, Institute of Automation of Chinese Academy of Sciences\textsuperscript{\(\diamond\)} \\
    Department of Computer \& Information Sciences, University of Strathclyde\textsuperscript{\(\S\)} \\
    \{jiaxin.zhang, moshfeghi.yashar\}@strath.ac.uk\textsuperscript{\(\S\)},  \\
    \{lizhongzhi2022, zhangmingliang2018\}@ia.ac.cn\textsuperscript{\(\diamond\)}, \\
    \{fyin, liucl\}@nlpr.ia.ac.cn\textsuperscript{\(\circledast\)}
}
\begin{document}
\maketitle
\begin{abstract}
Recent advancements in large language models (LLMs) and multi-modal models (MMs) have demonstrated their remarkable capabilities in problem-solving. Yet, their proficiency in tackling geometry math problems, which necessitates an integrated understanding of both textual and visual information, has not been thoroughly evaluated. To address this gap, we introduce the GeoEval benchmark, a comprehensive collection that includes a main subset of 2,000 problems, a 750 problems subset focusing on backward reasoning, an augmented subset of 2,000 problems, and a hard subset of 300 problems. This benchmark facilitates a deeper investigation into the performance of LLMs and MMs in solving geometry math problems. Our evaluation of ten LLMs and MMs across these varied subsets reveals that the WizardMath model excels, achieving a 55.67\% accuracy rate on the main subset but only a 6.00\% accuracy on the hard subset. This highlights the critical need for testing models against datasets on which they have not been pre-trained. Additionally, our findings indicate that GPT-series models perform more effectively on problems they have rephrased, suggesting a promising method for enhancing model capabilities.\footnote{The code and data available are at \url{https://github.com/GeoEval/GeoEval}.}
\end{abstract}

\section{Introduction}

\begin{figure}[tbhp!]
\centering
\includegraphics[width=0.45\textwidth]{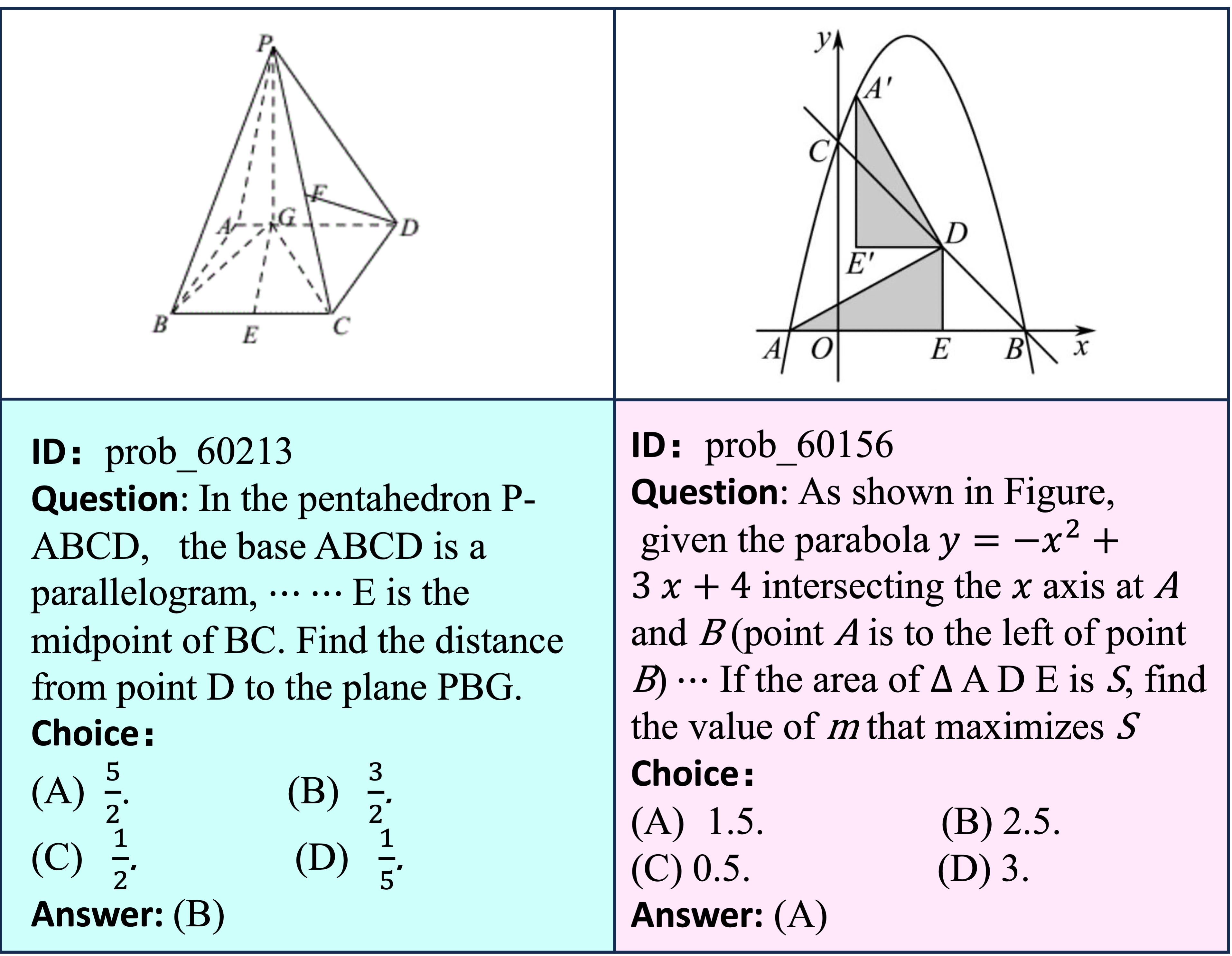}
\caption{Examples of the GeoEval benchmark.}
\label{fig:geoeval-hard-example}
\end{figure}

Geometry math problems are a key component in assessing the mathematical reasoning skills of K12 students, serving as a critical benchmark for evaluating educational outcomes \citep{GAOKAO_Benchmark}. The complexity of solving these problems stems from the requirement to interpret both textual and visual information, in addition to applying mathematical reasoning skills. This complexity has made geometry problem-solving a key area of interest for researchers aiming to evaluate the capabilities of AI models in this domain \citep{old-3, old-4, LANS, Solving-olympiad-geometry-without-human-demonstrations, GAPS, GOLD}.

In recent years, several datasets, such as Geometry3K \citep{geometry3k}, PGPS9K \citep{pgps9k}, and GeomVerse \citep{GeomVerse}, have been developed to test the proficiency of AI models in solving geometry math problems. Yet, these datasets often lack a standardized format and sufficient diversity, complicating the assessment of models' genuine proficiency in geometry problem-solving. Furthermore, these datasets typically focus on one type of geometry problem, such as flat geometry, overlooking other crucial areas like solid geometry. This oversight limits the ability to conduct a thorough evaluation across the full spectrum of geometry problems.

Simultaneously, advancements in large language models (LLMs) and multi-modal models (MMs) have demonstrated significant potential in handling complex reasoning tasks \citep{Program-of-Thoughts-Prompting, Chain-of-Thought, Multimodal-Chain-of-Thought, vlp}. This potential has raised considerable interest in testing these advanced models across a variety of tasks, such as math word problem solving \citep{MathVista} and physical problem solving \citep{ARB}. Despite this interest, specific research on evaluating these models' effectiveness in geometry problem-solving remains scarce. Therefore, it is critical to develop a new, comprehensive benchmark that can effectively assess LLMs and MMs in geometry problem-solving, especially considering the potential exposure of existing public datasets during model training \cite{NLP-Evaluation-in-trouble}. Comparing the performance of current LLMs and MMs on such a benchmark is essential, as it could yield valuable insights that further the development of models capable of tackling complex reasoning tasks.

To prompt research towards assessing LLMs' and MMs' proficiency in geometry math problem-solving, we introduce the GeoEval benchmark, a comprehensive collection specifically designed for this task. GeoEval features its \textit{Comprehensive Variety}, sourced from seven public datasets and formatted uniformly to encompass a wide range of geometric shapes. It includes \textit{Varied Problems}, covering flat, solid, and analytic geometry to challenge models comprehensively. GeoEval supports \textit{Dual Inputs}, accommodating both geometric diagrams and textual problem statements, making it suitable for evaluating both LLMs and MMs. To counter the potential overfitting to previously seen datasets, GeoEval introduces \textit{Diverse Challenges} through backward reasoning, augmented, and hard subsets, each designed to test different aspects of models' geometry problem-solving abilities. Additionally, GeoEval is annotated with \textit{Complexity Ratings}, allowing for a fine-grained analysis of model performance across various difficulty levels, thus providing a robust framework for advancing AI capabilities in understanding and solving geometry math problems. Examples of geometry problems from our GeoEval can be found in Figure~\ref{fig:geoeval-hard-example}.

In this paper, we conduct extensive experiments using the GeoEval benchmark to evaluate the proficiency of ten LLMs and MMs in solving geometry problems. This includes three LLMs: CodeGen2-16B \citep{CodeGen2}, GPT-3.5 \citep{ChatGPT}, and GPT-4 \citep{GPT-4}; two LLMs specialized in mathematics: WizardMath-70B and WizardMath-7B-V1.1 \citep{WizardMath}; and five MMs: llava-7B-V1.5 \citep{llava}, Qwen-VL \citep{Qwen-VL}, mPLUG-Owl2 \citep{mPLUG-Owl2}, InstructBLIP \citep{InstructBLIP}, and GPT-4V \cite{GPT-4}. The findings reveal that GeoEval forms a challenging benchmark, with both LLMs and MMs struggling to resolve its complexities effectively. 

Notably, our results indicate that: \circled{1} Models pre-trained on mathematical corpora, such as the WizardMath models, deliver superior performance across various GeoEval subsets (Section~\ref{sec:comparison-among-text-only-llms}), establishing new benchmarks in the field. \circled{2} One advantage of these models is that they implicitly encompass the required mathematical knowledge demanded to solve geometry math problems (Section~\ref{sec:external-knowledge-required}).  \circled{3} However, we also find that though pre-training on a mathematical corpus is crucial for solving geometry math problems, it may not be enough (Section~\ref{sec:analysis-on-the-best-model}). \circled{4} Additionally, we observe that GPT series models exhibit enhanced problem-solving efficiency when tackling geometry questions that they have previously rephrased (Section~\ref{sec:analysis-on-the-best-model}). \circled{5} Further analyses underscore the value of incorporating descriptions of geometric diagrams, which significantly aids LLMs in understanding and solving geometry problems (Section~\ref{sec:addition-help-from-the-geometric-diagram-descriptions}). \circled{6} Finally, our experiments show that the performance of both LLMs and MMs declines as the problem length and complexity of the problem increases (Section~\ref{sec:different-length}). Through the GeoEval benchmark, we believe this research provides the first comprehensive quantitative assessment of the latest LLMs and MMs in the domain of geometry problem-solving.

\section{Related Work}

Numerous benchmarks have been developed to assess the capabilities of LLMs in the the geometry problem-solving task. However, these benchmarks face limitations, such as restricted access, like GEOS \citep{GEOOS} and GeoShader \citep{GeoShader} datasets, or insufficient scale, as seen with GEOS++ \citep{GEOOS++}. Although recent efforts have introduced new benchmarks like Geometry3K \citep{geometry3k}, UniGeo \citep{unigeo}, and PGPS9K \citep{pgps9k}, they still fall short in offering a uniform format and embracing a wide range of problem types. In response, we introduce the comprehensive and challenging GeoEval benchmark, aiming to advance the evaluation of geometry problem-solving abilities.

Recently, LLMs \citep{Vicuna, LLaMA, ChatGPT} and MMs \citep{llava, mPLUG-Owl2, GPT-4} have achieved impressive results on complex tasks, attracting research into their performance across specialized tasks. Previous work like MathVista \citep{MathVista} have concentrated on scientific domains, likewise SEED \citep{SEED} explores models' understanding of temporal and spatial relationships. Despite these advancements, there remains a gap in the examination of models' ability to solve geometry math problems. Through the GeoEval benchmark, we aim to fill this gap by offering a detailed assessment of both LLMs' and MMs' abilities to tackle a variety of geometry math challenges.

\section{GeoEval Dataset}

The GeoEval benchmark is structured into four subsets: GeoEval-2000, comprising 2,000 problems; GeoEval-backward, with 750 problems; GeoEval-aug, containing 2,000 problems; and GeoEval-hard, including 300 problems. The subsequent sections will detail the collection process for each subset, followed by an explanation of the unique features of the GeoEval benchmark.\footnote{Statistics for the GeoEval benchmark are in Appendix~\ref{sec:appendix-statistical-analysis}.}

\begin{table*}[tbhp!]
\centering
\begin{small}
\begin{tabular}{@{}lccccc@{}}
\toprule
\textbf{Dataset} &
  \begin{tabular}[c]{@{}c@{}}\textit{Comprehensive} \\ \textit{Variety}\end{tabular} &
  \begin{tabular}[c]{@{}c@{}}\textit{Varied} \\ \textit{Problems}\end{tabular} &
  \begin{tabular}[c]{@{}c@{}}\textit{Dual} \\ \textit{Inputs}\end{tabular} &
  \begin{tabular}[c]{@{}c@{}}\textit{Diverse} \\ \textit{Challenges}\end{tabular} &
  \begin{tabular}[c]{@{}c@{}}\textit{Complexity} \\ \textit{Ratings}\end{tabular} \\ \midrule
MathQA \citep{MathQA}     & n/a   & flat       & text &  \xmark & \xmark \\
GeometryQA \citep{geometryqa} & n/a   & flat       & text &  \xmark & \xmark \\
Geometry3K \citep{geometry3k} & n/a   & flat       & text + diagram & \xmark & \xmark \\
GeoQA+ \citep{geoqa+}    & n/a   & flat       & text + diagram & \xmark & \xmark \\
MATH \citep{MATH}       & n/a   & flat       & text & \xmark & \xmark \\
UniGeo \citep{unigeo}    & n/a   & flat       & text + diagram & \xmark & \xmark \\
PGPS9K \citep{pgps9k}    & n/a   & flat       & text + diagram & \xmark & \xmark \\ 
GeomVerse \citep{GeomVerse}  & n/a  & flat       & text + diagram & \xmark & \cmark \\
\midrule
MathVista \citep{MathVista}  & 4  & flat       & text + diagram & \xmark \(^\ddagger\)  & \xmark \\
GeoEval    & 7 + 3 (new)  & flat, solid, analytic & text + diagram & \cmark & \cmark \\ \bottomrule
\end{tabular}
\end{small}
\caption{Comparison between GeoEval benchmark and other datasets. Under \textit{Comprehensive Variety}, MathVista and GeoEval stand out as collective datasets, while the rest, are denoted as 'n/a'. GeoEval includes problems from seven public datasets and three newly created ones. \textit{Varied Problems} categorizes problems into "flat geometry", "solid geometry", and "analytic geometry", For \textit{Dual Inputs}, "text" signifies problems presented only in text format, whereas "text + diagram" encompasses problems with both texts and diagrams. In \textit{Diverse Challenges}, the symbol \(\ddagger\) indicates that MathVista introduces three new datasets, which, however, are unrelated to the geometry problem-solving task.}
\label{tab:comparison-with-previous-datasets}
\end{table*}

\subsection{Data Collection}

\subsubsection{Collection from Diverse Data Sources}

We have compiled a comprehensive collection of public geometry math problem datasets, with a total of 24,912 geometry math problems from sources such as Geometry3K \cite{geometry3k}, PGPS9K \citep{pgps9k}, UniGeo \citep{unigeo}, GeoQA+ \citep{geoqa+}, GeometryQA \citep{geometryqa}, as well as geometry problems from the MATH \citep{MATH} and MathQA \citep{MathQA} datasets. The first four datasets feature geometry questions that include both problem texts and geometric diagrams, whereas the latter three datasets comprise questions that only contain problem texts. Detailed information about all source datasets is available in Appendix~\ref{sec:appendix-source-datasets}.

Building on the data gathered, we then selected 2,000 geometry math problems to create our GeoEval-2000 subset. This selection process was guided by the aim to inclusively cover a wide range of basic geometric shapes, ensuring a broad representation of geometry concepts. The distribution of geometric shapes within this subset is further detailed in Appendix~\ref{sec:appendix-distributions-of-different-geometric-shapes}.

\subsubsection{Backward Data Generation}

In contrast to forward problems, backward problems use the answer from forward problems as a starting point, posing a query to determine a specific number that was part of the forward problems but is concealed in the backward problems \citep{Forward-Backward-Reasoning}. These types of questions are particularly effective in assessing models' capability for multi-step reasoning. Following the methodology of previous research \citep{MetaMath}, we selected 750 problems from the GeoEval-2000 subset and created corresponding backward questions. This process involved masking a number, the solution of the forward problems, as "X". The prompt "\textit{The correct answer is \(\text{ans}_{\text{gold}}\). Now please answer what is the value of X?}", where \(\text{ans}_{\text{gold}}\) represents the correct answer to the forward problems, is then added. The example of backward problems can be found in Appendix~\ref{sec:appendix-backward-question-example}.

\subsubsection{Augmented Data Generation}

To evaluate the resilience of current models and mitigate the risk of data leakage that may occur during the pre-training phase, we implement a context learning strategy for rephrasing problems from the GeoEval-2000 subset. Each problem is rephrased into five variant candidates by GPT-3.5 \cite{ChatGPT}, ensuring they retain the original problem's semantic essence while varying in lexical structure. Out of these five alternatives, one is selected randomly to substitute the original problems, forming the GeoEval-aug subset.

\subsubsection{Hard Data Collection}

While the GeoEval-2000 subset comprises geometry problems from a variety of source datasets, it exhibits a lack of diversity in problem categories, notably in solid geometry and analytic geometry. To enhance the diversity of problem categories, we introduce the GeoEval-hard subset, which includes 300 geometry problems specifically focusing on solid geometry and analytic geometry, providing a broader assessment scope. More details regarding the comparison between the GeoEval-hard subset with other datasets are in Appendix~\ref{sec:appendix-comparison-between-geohard-subset-and-other-public-datasets}.

The GeoEval-hard subset sources from the copyrighted collection containing 10,000 geometry math problems created based on templates summarized from the online resources. An initial selection is made using a rule-based engine equipped with a keyword list, targeting solid and analytic geometry problems. This step yields around 3,100 potential problems, identified as the GeoEval-hard-raw subset. Next, a manual review further narrows these down to 300 problems related to solid and analytic geometry. The cleaning and manual inspection process is documented in Appendix~\ref{sec:appendix-data-correctness-check-for-geoeval-hard}.

\subsection{Features of GeoEval}

The GeoEval benchmark is specifically designed to assess the ability of models to resolve geometric math problems. This benchmark features five characteristics: \textit{Comprehensive Variety}, \textit{Varied Problems}, \textit{Dual Inputs}, \textit{Diverse Challenges}, and \textit{Complexity Ratings}, with each attribute exemplified in the Appendix~\ref{sec:appendix-examples-from-geoeval-representing-five-features}. For an insightful contrast, Table~\ref{tab:comparison-with-previous-datasets} offers a comparative analysis of GeoEval against earlier datasets.

\paragraph{\textit{Comprehensive Variety}}
GeoEval consists of a diverse collection of geometry problems sourced from the seven most recent datasets. Therefore, the problems in GeoEval cover a wide range of geometric shapes, offering a comprehensive view of varied geometry math challenges.

\paragraph{\textit{Varied Problems}}
The GeoEval benchmark encompasses three distinct categories of geometry math problems, namely flat geometry, solid geometry, and analytic geometry.

\paragraph{\textit{Dual Inputs}}
GeoEval features problems in two formats: those accompanied by diagrams and those consisting solely of text. This versatility makes it suitable for evaluating models that process diagrams or text-based inputs.

\paragraph{\textit{Diverse Challenges}}
In addition to gathering public datasets, GeoEval also generates its out-of-distribution dataset aimed at addressing data leakage problems. This includes a backward reasoning subset, an augmented subset, and a hard subset, all created by us.

\paragraph{\textit{Complexity Ratings}}
GeoEval is equipped with annotations indicating the complexity level for each problem, serving as a guideline to evaluate models' proficiency in solving these tasks.\footnote{Algorithm for classifying complexity is in Appendix~\ref{sec:appendix-complexity}.}

\section{Experiments}

\subsection{Experimental Setup}
\label{sec:experimental-setup}

In this study, we deliberately select state-of-the-art LLMs and MMs that are widely recognized for their advanced capabilities, including:

\begin{itemize}
    \item{\textbf{LLMs Specialized in Programming Code}}: We include CodeGen2-16B model \citep{CodeGen2}, which is renowned for its proficiency in understanding and generating programming code, offering insights into its adaptability to solve geometry math problems.
    \item{\textbf{LLMs with a Focus on Mathematics}}: This includes WizardMath-7B-V1.1 and WizardMath-70B  \citep{WizardMath}, explicitly pre-trained on mathematical corpora. Their inclusion allows for an assessment of models that have been fine-tuned to tackle complex mathematical problems.
    \item{\textbf{LLMs Designed for a Broad Range of Topics}}: Models such as GPT-3.5 \citep{ChatGPT} and GPT-4 \citep{GPT-4} exemplify the advanced commercial LLMs engineered to encompass a broad range of topics.
    \item{\textbf{Multi-Modal Models (MMs) with Diverse Decoders}}: Given the ubiquity of ViT architecture \citep{ViT} as the vision encoder in MMs, we select models that integrate ViT with various LLMs as decoders. This includes llava-7B-V1.5 \citep{llava} with Vicuna \citep{Vicuna}, Qwen-VL \citep{Qwen-VL} using Qwen \citep{qwen}, mPLUG-Owl2 \citep{mPLUG-Owl2} with LLaMA \citep{LLaMA}, InstructBLIP \citep{InstructBLIP} with Vicuna \citep{Vicuna}, and GPT-4V \citep{GPT-4}.
\end{itemize}

These models are evaluated through a zero-shot approach, utilizing straightforward instruction prompts to directly assess their geometry problem-solving capabilities without further fine-tuning specifically for our benchmark.\footnote{Details on the prompt design and the hyper-parameters used for these models are available in Appendix~\ref{sec:appendix-evaluation-details}.} 

\subsection{Evaluation Metric}
\label{sec:evaluation-metric}

\begin{table*}[tbhp!]
\centering
\begin{tabular}{@{}l|cc|c|c|c@{}}
\toprule
                   & \multicolumn{2}{c|}{GeoEval-2000} & GeoEval-backward & GeoEval-aug & GeoEval-hard \\ 
Model              & A (\%)      & T (\%)      & A (\%)       & A (\%)  & A (\%)   \\ \midrule
CodeGen2-16B $\lozenge$     & 28.76           & 22.06           & 5.10            & 8.50        & 5.66         \\
GPT-3.5 $\lozenge$           & 24.71           & 21.27           & 22.66            & 41.25       & 22.33        \\
GPT-4 $\lozenge$             & 27.95           & 43.86           & 26.00            & 45.75       & 10.10        \\ \midrule
WizardMath-70B $\lozenge$    & \textbf{55.67}            & 34.20           & 28.66            & 37.75       & 6.00         \\
WizardMath-7B-V1.1 $\lozenge$ & 54.78           & 32.76           & 32.66           &  \textbf{47.75}       & 6.00         \\ \midrule
llava-7B-V1.5      & 12.80           & 21.01           & 11.33            & 20.25       & 20.30         \\
Qwen-VL            & 25.60           & 25.97         & 5.66            & 22.25       & 21.66        \\
mPLUG-Owl2         & 37.76           & n/a          & \textbf{35.33}           & 38.00       &  \textbf{22.66}        \\
InstructBLIP $\dagger$      & 52.18         & n/a        & 15.66           & 35.00      & 70.30       \\
GPT-4V             & 37.22           & \textbf{43.86} $\ddagger$           & 26.00            & 45.75       & 10.10        \\ \bottomrule
\end{tabular}
\caption{Accuracy scores of models on our GeoEval benchmark. The "$\lozenge$" refers to all LLMs. The "A" signifies the overall accuracy across all problems, while "T" denotes the accuracy for problems containing only texts without diagrams. The "n/a" indicates that scores are unavailable due to models cannot process text-only inputs. The "$\dagger$" shows our doubt on the high accuracy rates reported by the IntructBLIP model, our point is elaborated in Section~\ref{sec:experimental-results}. The "$\ddagger$" notes that the accuracy figures for GPT-4V are derived from GPT-4, as GPT-4V does not support image-free inputs. Detailed reporting on model performance, segmented by dataset origins, is available in Appendix~\ref{sec:appendix-data-sources}.}
\label{tab:overall-results}
\end{table*}

Building upon the approach by MathVista \citep{MathVista}, we first input the generated sequence from the model into GPT-4 to extract the target value or option letter. To enhance the precision of our answer extraction, we formulate intricate rules for post-processing the outcomes in cases where GPT-4 falls short. Specifically, our extraction pipeline involves two steps: firstly, using a prompt to extract the answer. Secondly, employing regular expressions to extract any remaining answers that couldn't be obtained from the first step. Please refer to Table \ref{tab:promt_answer_extraction} and Table \ref{tab: human_ext_rules} in Appendix for the extraction instruction and the constructed samples. This approach has enabled us to attain an extraction accuracy surpassing 97\%\footnote{We assess the accuracy of the extraction by manually reviewing 200 uniformly sampled examples.}, similar to the success rate reported in MathVista \citep{MathVista}.

The extracted results are compared against the golden answers to determine the final performance metric. Given the model's intention to produce responses in varying formats, either as the precise answer (for instance, "3.15") or as the corresponding option letter (such as "A"), we regard a prediction as accurate if it either matches the golden answer or the golden option letter.

\subsection{Experimental Results}
\label{sec:experimental-results}

In this section, we present the accuracy achieved by models on our GeoEval benchmark. Table~\ref{tab:overall-results} highlights that models pre-trained on a math-specific corpus tend to outperform others. Furthermore, except for llava-7B-V1.5 and Qwen-VL, multi-modal models (MMs) generally exceed the performance of large language models (LLMs). Notably, InstructBLIP exhibits exceptionally high accuracy scores across all subsets, yet its results raise some concerns, and we have chosen to exclude the InstructBLIP model. The rationale behind this decision is detailed in Appendix~\ref{sec:appendix-reason-for-removing-instructblip}.

\subsubsection{Comparison among LLMs}
\label{sec:comparison-among-text-only-llms}

When reviewing the performances of LLMs as detailed in Table~\ref{tab:overall-results}, it becomes evident that models pre-trained on mathematical corpora demonstrate superior efficacy in solving geometry math problems compared to those trained on general corpora. Specifically, evaluating on all problems of the GeoEval-2000 subset (marked as "A" in the table), WizardMath-70B leads with an accuracy of 55.67\%, while WizardMath-7B-V1.1 closely follows with a 54.78\% accuracy, outperforming other LLMs. Conversely, GPT-4, GPT-3.5, and CodeGen2-16B report notably lower accuracies, all under 30.00\%. Focusing on questions solely based on problem text within the GeoEval-2000 subset (indicated as "T" in the table), GPT-4 emerges as the frontrunner, securing the highest accuracy of 43.86\%, with WizardMath models also surpassing the 32.00\% accuracy. These findings underscore the enhanced proficiency of models pre-trained on math-specific corpora in tackling geometry math problems, particularly when problems are well-described textually, as evidenced by GPT-4's leading performance.

In the GeoEval-backward subset, WizardMath-7B-V1.1 excels with the highest accuracy of 32.66\%, closely followed by WizardMath-70B at 28.66\%. This significant drop in performance across all LLMs, compared to the GeoEval-2000 results, highlights a collective weakness in backward reasoning capabilities. For the GeoEval-aug subset, WizardMath-7B-V1.1 again tops the leaderboard with an accuracy of 47.75\%, with GPT-4 not far behind at 45.75\% accuracy. Lastly, within the GeoEval-hard subset, all models, excluding GPT-3.5, exhibit relatively low accuracies, indicating a broad difficulty in addressing the most challenging solid geometry and analytic geometry problems. To investigate the reason for GPT-3.5 achieves better performance than GPT-4 on the GeoEval-hard subset, we find that GPT-4 tends to generate verbose solutions, often accompanied by code, which causes it to either terminate before solving the problem or enter a self-cycling loop of generating redundant information, failing to provide a final answer. In contrast, GPT-3.5 adopts a more concise approach, consistently producing option letters (e.g., "A") following the reasoning steps as solutions. We believe this concise solution generation strategy contributes to GPT-3.5's relatively better performance on the GeoEval-hard subset.

\subsubsection{Comparison among Multi-Modal Models}

Table~\ref{tab:overall-results} shows that among the MMs, GPT-4V and mPLUG-Owl2 consistently outperform their counterparts across all subsets. Specifically, within the GeoEval-2000 subset, mPLUG-Owl2 leads with an accuracy of 37.76\%, closely followed by GPT-4V at 37.22\%, with the remaining MMs falling behind at lower accuracies. Specifically, Qwen-VL and llava-7B-V1.5 achieve accuracies of 25.60\% and 12.80\%, respectively. When examining problems that only involve texts, GPT-4V achieves a 43.86\% accuracy, significantly surpassing llava-7B-V1.5 (21.01\%) and Qwen-VL (25.97\%).

In the GeoEval-backward subset, mPLUG-Owl2 tops with the accuracy of 35.33\%, with GPT-4V following at 26.00\% accuracy. This performance shows a notable lack of backward reasoning skills, as illustrated by the diminished results of llava-7B-V1.5 and Qwen-VL in this category. Moving to the GeoEval-aug subset, GPT-4V leads with an impressive 45.75\% accuracy, with mPLUG-Owl2 in second place with 38.00\% accuracy. Both Qwen-VL and llava-7B-V1.5 show comparable performances in this subset. Lastly, within the GeoEval-hard subset, mPLUG-Owl2 demonstrates the highest efficacy with a 22.66\% accuracy, closely followed by Qwen-VL and llava-7B-V1.5. Surprisingly, GPT-4V records a lower accuracy of just 10.10\%, highlighting the challenging nature of the GeoEval-hard subset and the varied capabilities of MMs in addressing the most difficult problems.

\subsubsection{Comparison between LLMs and Multi-Modal Models}

In the GeoEval-2000 subset, specifically for problems that only include texts, GPT-4's performance exceeds the top MMs, Qwen-VL, by 17.89\%. This is attributed to the MMs' inability to access geometric diagrams, which likely hinders their comprehension of the problems. Moreover, when evaluating all problems of the GeoEval-2000 subset, WizardMath-70B surpasses the best MMs, Qwen-VL, by 17.91\% in accuracy. However, MMs like GPT-4V and mPLUG-Owl2 achieve significantly higher accuracy than LLMs not pre-trained on mathematical content. This underscores the value of mathematical pre-training for excelling in geometry problem-solving. Notably, GPT-4V's accuracy on all GeoEval-2000 problems is 9.27\% higher than GPT-4's, suggesting GPT-4V's superior capability in solving geometry problems with diagrams.

This pattern persists in the GeoEval-aug subset, where WizardMath-7B-V1.1, a model trained on a mathematical corpus, achieves the highest accuracy at 47.75\%. Conversely, mPLUG-Owl2 leads in the GeoEval-backward and GeoEval-hard subsets, with accuracies of 35.33\% and 22.66\%, respectively. Given that GeoEval-aug rephrases questions from GeoEval-2000, it implies both subsets might have been exposed to the models during their pre-training phase. In contrast, GeoEval-backward and GeoEval-hard subsets are less likely to have been previously exposed. This suggests that WizardMath-7B-V1.1 excels with familiar geometry math problems, while mPLUG-Owl2 demonstrates a robust capability in tackling unseen geometry problems. This is further evidenced by the low performance of WizardMath models on the GeoEval-hard subset, where both models only achieve an accuracy of 6.00\%.

\subsubsection{Analysis on the Best Model}
\label{sec:analysis-on-the-best-model}

Table~\ref{tab:overall-results} shows that GPT-4, the leading LLMs, records the highest accuracy on the GeoEval-aug subsets, though it only secures a 27.95\% accuracy on the GeoEval-2000 subset. A similar pattern of improvement is noted for the GPT-3.5 model, which sees its accuracy jump from 24.71\% on the GeoEval-2000 subset to 41.25\% on the GeoEval-aug subset. This improvement aligns with the involvement of GPT-3.5 in generating the GeoEval-aug subset, suggesting that the capabilities of GPT-3.5 and GPT-4 in addressing geometry math problems significantly benefit from their use in rephrasing geometry question texts.

While WizardMath-70B and WizardMath-7B-V1.1, both pre-trained on a mathematical corpus, demonstrate superior performance on the GeoEval-2000 subset, they show a marked decline in accuracy across the other subsets, with the most significant decreases observed on the GeoEval-hard subset. This indicates that although pre-training on a mathematical corpus is crucial for solving geometry math problems, it may not be enough.

In contrast to the significant variances in accuracy observed among LLMs across different subsets, the top-performing multi-modal model, mPLUG-Owl2, maintains relatively stable accuracies with scores of 37.76\% on the GeoEval-2000, 35.33\% on the GeoEval-backward, and 38.00\% on the GeoEval-aug subsets. Additionally, the performance of GPT-4V on the GeoEval-aug subset surpasses its accuracy on the GeoEval-2000 subset, mirroring the trends observed with GPT-4 and GPT-3.5, further illustrating the enhanced effectiveness of GPT-series models when engaged in rephrasing the content of geometry questions.

\subsection{Results Across Different Subjects}

\begin{figure}[tbhp!]
\centering
\includegraphics[width=0.45\textwidth]{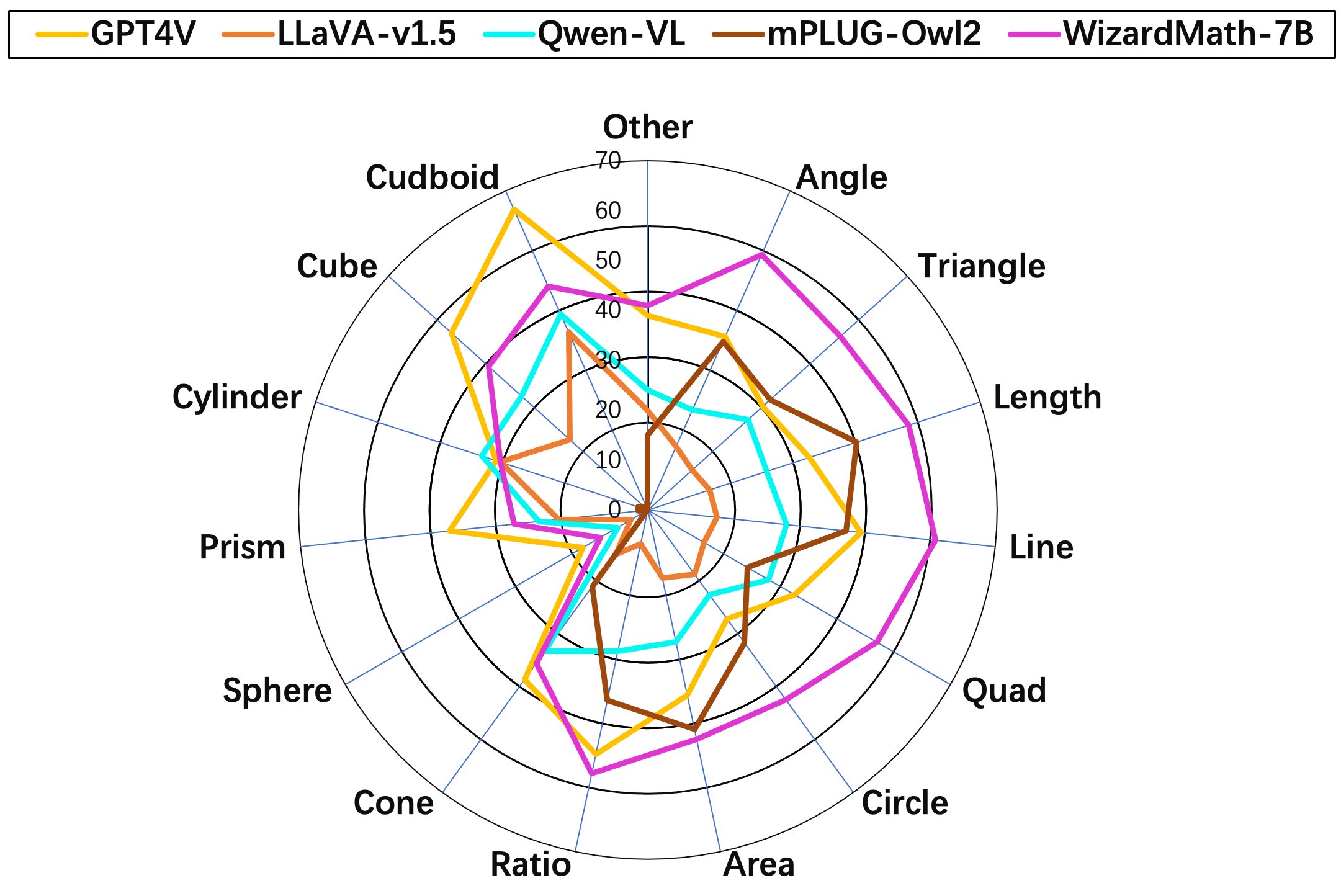}
\caption{Detailed accuracy scores for models across various academic subjects.}
\label{fig:different-subjects}
\end{figure}

Figure~\ref{fig:different-subjects} displays the performance of models across various subjects, revealing distinct strengths. The WizardMath-7B model significantly outperforms others in flat geometry problems, such as length and lines. Conversely, in solid geometry problems like cuboids and spheres, GPT-4V surpasses WizardMath-7B, indicating its superior capability in addressing solid geometry questions.

\subsection{Benefit from the Geometric Diagram Descriptions}
\label{sec:addition-help-from-the-geometric-diagram-descriptions}

\begin{table}[tbhp!]
\centering
\begin{small}
\begin{tabular}{@{}lcc@{}}
\toprule
Models        & \multicolumn{1}{c}{\xmark} & \multicolumn{1}{c}{\cmark} \\ \midrule
GPT-4V        & 40.28                & 45.61 (\textit{+5.33})               \\
WizardMath-7B & 38.10                & 56.83 (\textit{+18.73})               \\ \bottomrule
\end{tabular}
\caption{Comparison of models with (\cmark) and without (\xmark) geometric diagram descriptions.}
\label{tab:diagram-descriptions}
\end{small}
\end{table}

To assess the impact of including geometric diagram descriptions on models' ability to comprehend geometric diagrams and solve related problems, we selected a sample of 300 questions with geometric diagram descriptions from the GeoEval-2000 subset. We then evaluated the performance of two models, GPT-4V and WizardMath-7B-V1.1, on these questions, both with and without the use of geometric diagram descriptions, which describe the geometric shapes and relations encapsulated in the diagram. The results in Table~\ref{tab:diagram-descriptions} indicate that GPT-4V's accuracy decreases by 5.33\% without the diagram descriptions. More significantly, WizardMath-7B's accuracy falls by 18.73\% in the absence of these descriptions. This evidence suggests that supplemental geometric diagram descriptions significantly enhance models' efficiency in solving geometry math problems, particularly benefiting LLMs.

\subsection{External Constants Required for Solving the Problems}
\label{sec:external-knowledge-required}

\begin{figure}[tbhp!]
\centering
\includegraphics[width=0.47\textwidth]{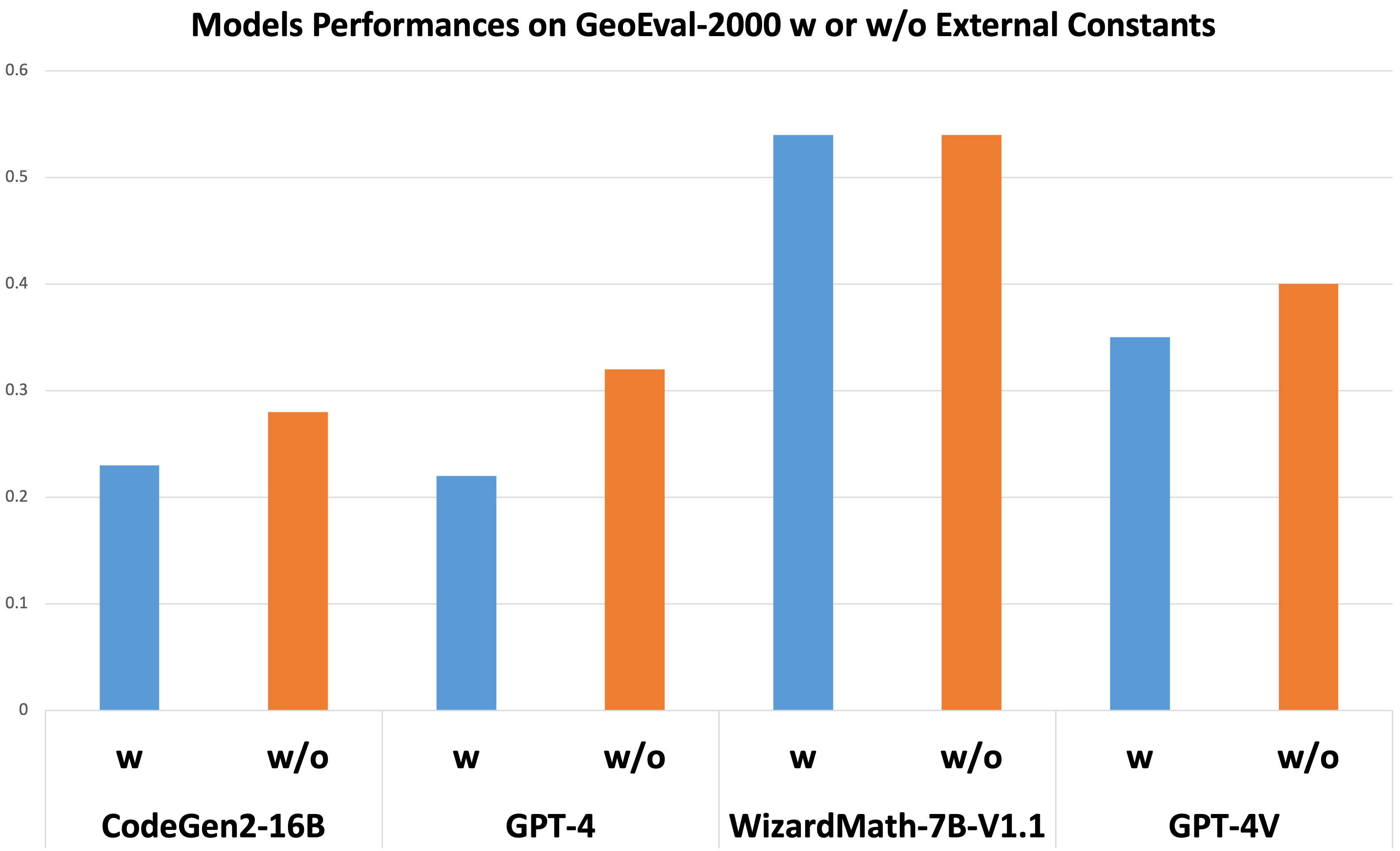}
\caption{Comparison of models requiring external constants ("w" in blue color) and those do not ("w/o" in orange color).}
\label{fig:external-knowledge}
\end{figure}

In the GeoEval benchmark, certain questions require external constants, such as the value of \(\pi\), which is not typically included in the problem text. This necessitates models to have pre-existing knowledge to accurately solve these problems. Figure~\ref{fig:external-knowledge} assesses the performance of four models on problems differentiated by the need for external constants, identified through a heuristic approach that classifies problems according to whether their solutions require constants.

Figure~\ref{fig:external-knowledge} shows that the WizardMath-7B-V1.1 model maintains consistent accuracy on the GeoEval-2000 subset, regardless of the requirement for external constants, unlike other models, which perform better on problems without such requirements. This consistency in WizardMath-7B-V1.1's performance is likely due to its pre-training on a math-specific corpus, providing it with the necessary knowledge to resolve geometry math problems effectively. In contrast, models trained on general corpora may not possess this specialized mathematical knowledge, hindering them from using external constants to solve the problems correctly.

% Further analysis on the GeoEval-aug dataset, examining performance on problems with and without the need for external knowledge, reveals that WizardMath-7B-V1.1's accuracy slightly declines on the GeoEval-aug compared to GeoEval-2000, yet it remains consistent for problems not requiring external knowledge. This pattern of consistent also exists in GPT-4V and GPT-4 models, suggesting a general robustness in these models' ability to process different representations of problem texts.

\subsection{Performances According to Different Problem Lengths and Varied Complexities}
\label{sec:different-length}

\begin{figure}[tbhp!]
\centering
\includegraphics[width=0.48\textwidth]{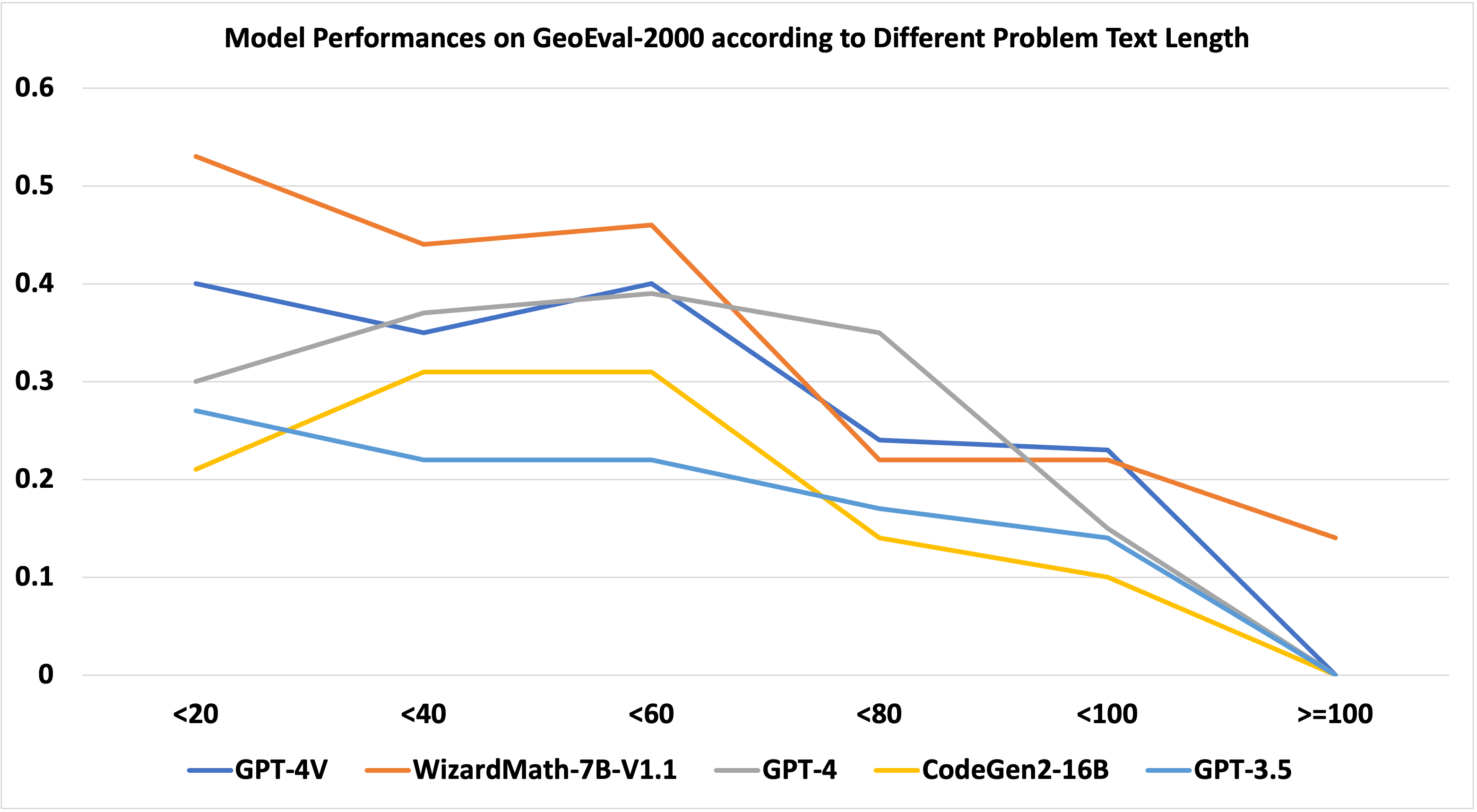}
\caption{Models performances on GeoEval-2000 subset according to different question lengths.}
\label{fig:problem-length}
\end{figure}

Figure~\ref{fig:problem-length} shows how models perform with inputs of different lengths. Performance slightly varies for problems ranging from 80 to 100 characters, but there's a clear trend of decreasing accuracy as problem length increases. This is expected, as longer questions typically involve more complex geometry math problems, challenging the models more as the length grows. The figure also points out that the WizardMath-7B-V1.1 model is notably more adept at handling longer questions, with GPT-4V and GPT-4 showing relatively stable accuracy for increased question lengths. On the other hand, GPT-3.5 and CodeGen2-16B perform less effectively on lengthy questions.

\begin{figure}[tbhp!]
\centering
\includegraphics[width=0.48\textwidth]{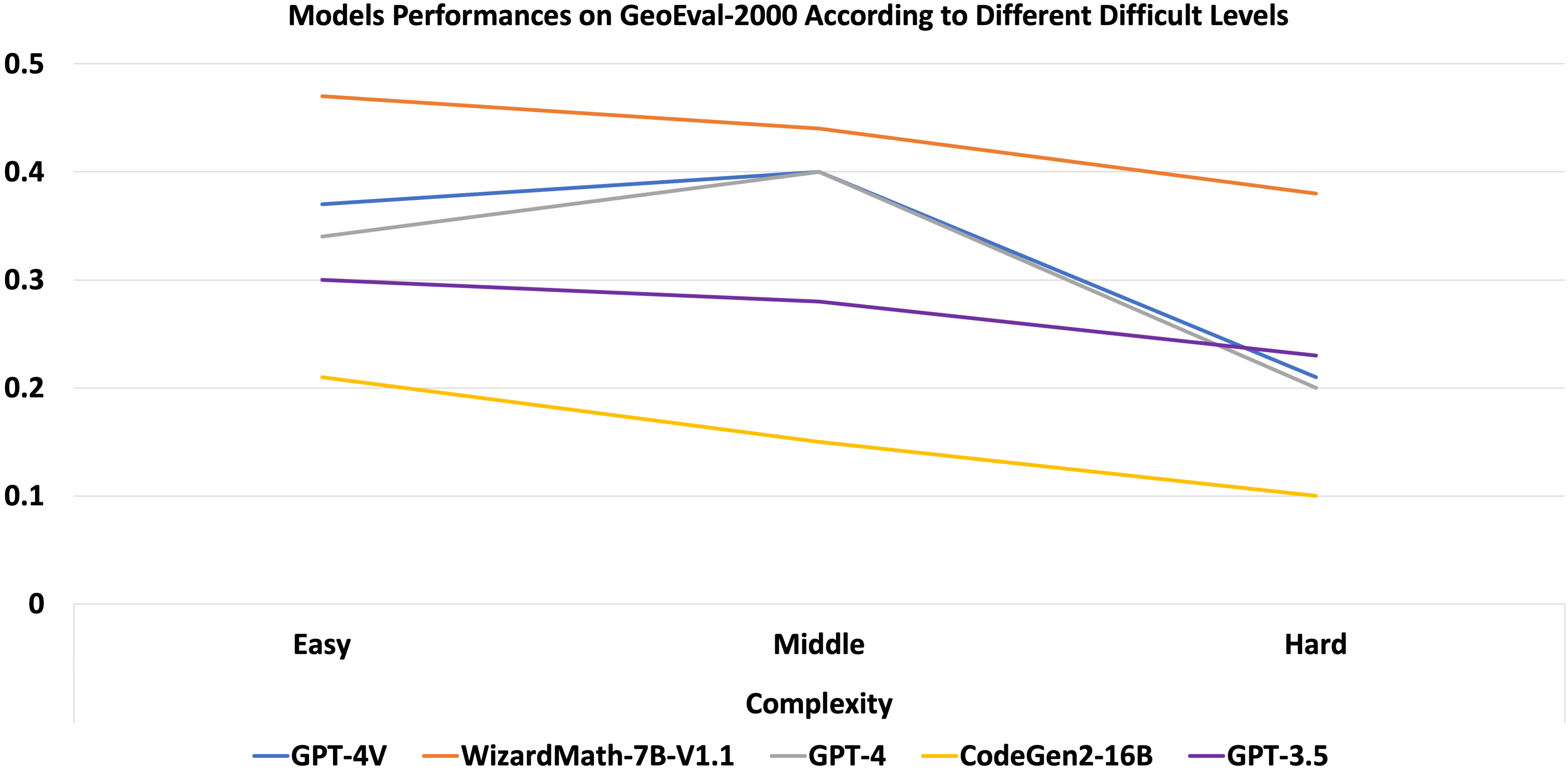}
\caption{Model performances on GeoEval-2000 subset according to different complexity levels.}
\label{fig:complexity-2}
\end{figure}

Upon the analysis in Figure~\ref{fig:complexity-2}, similar to the observations made in Figure~\ref{fig:problem-length} regarding input lengths, we delve into the models' performances as they relate to the complexity of geometry math problems. Figure~\ref{fig:complexity-2} presents the performance of models across varying levels of problem complexity. It is evident that as the complexity of geometry problems escalates, the accuracy of the models correspondingly diminishes.

\section{Conclusion}

In this study, we present GeoEval, a benchmark developed to assess the geometry problem-solving capabilities of large language models (LLMs) and multi-modal models (MMs). GeoEval comprises four distinct subsets, each designed to facilitate a thorough evaluation. Through our assessment of ten cutting-edge LLMs and MMs using the GeoEval benchmark, we underscore the critical role of mathematical corpus pre-training for effective geometry problem resolution. This is exemplified by the WizardMath model's leading performance on the GeoEval-2000 subset, achieving an accuracy of 55.67\%. However, the WizardMath model's challenges with the GeoEval-hard subset suggest a need for enhanced reasoning skills. Additionally, our analysis reveals that GPT-series models exhibit improved performance on geometry problems they have rephrased, pointing to the potential benefits of self-rephrasing in problem-solving.

\section{Limitations}

This study, while providing significant insights into the capabilities of large language models (LLMs) and multi-modal models (MMs) in solving geometry problems, has several limitations. 

One primary constraint is that our evaluation predominantly focuses on quantitative metrics of accuracy, potentially overlooking qualitative aspects of model reasoning and explanation that are crucial for educational applications. The performance of models on the hard subset also highlights a gap in advanced reasoning abilities, suggesting that current LLMs and MMs, including those pre-trained on mathematical corpora, may still struggle with highly complex or novel problem types. In addition, we conduct experiments focusing on testing the models' ability to recall and effectively utilize knowledge about mathematical constants. However, a comprehensive evaluation of external knowledge utilization is required, such as theorems and principles, which are beyond just mathematical constants. We plan to explore models' abilities to leverage diverse forms of external knowledge in future work.

Moreover, this work reveals the effectiveness of rephrased problems by GPT-series models and suggests a specific interaction effect that may not generalize across all types of geometry problems or other LLMs and MMs, indicating a need for broader research to fully understand the implications of rephrasing on model performance.

\section*{Acknowledgments}

This work has been supported by the National Key Research and Development Program Grant 2020AAA0109700, and the National Natural Science Foundation of China (NSFC) Grant U23B2029.

\bibliography{custom}
\bibliographystyle{acl_natbib}

\appendix

\begin{table*}[tbhp!]
\centering
\small
\renewcommand\tabcolsep{3.0pt} % column space
\begin{tabular}{cp{12cm}}
 \toprule
 \textbf{Element} & \textbf{Prompt} \\
 \midrule
 Task description & 
 \begin{minipage}[s][0.6cm]{1.5\columnwidth}
 You are a result extraction bot. I will provide you with geometry questions and a model output, and you will help me extract the reference answers from the model's output.
 \end{minipage}
 \\
 \midrule
 Example 1 & 
 \begin{minipage}[s][3.0cm]{1.5\columnwidth}
 \textbf{Question:} As shown in the figure, in triangle $ABC$, $AB=AC$, $\angle A=40^\circ$, $DE$ is the perpendicular bisector of $AB$. What is the degree measure of $\angle DBC$? ()\\
\textbf{Choices:} (A) 30.0 (B) 20.0 (C) 15.0 (D) 10.0\\
 \textbf{Model response:} Given that $\angle A = 40^\circ$, $AB = AC$, we can conclude that $\angle ABC = \angle ACB = 70^\circ$. Also, since $DE$ is perpendicular and bisects $AB$, we have $DB = AD$ and $\angle ABD = \angle A = 40^\circ$. Therefore, $\angle DBC = \angle ABC - \angle ABD = 70^\circ - 40^\circ = 30^\circ$. The answer is option A.
\\
 \textbf{Extracted answer:} \textbf{A}\\
 \end{minipage}
 \\
 \midrule
 Example 2 & 
 \begin{minipage}[s][3.0cm]{1.5\columnwidth}
 \textbf{Question:} As shown in the diagram, the lateral surface area of a cone with a radius of the base $r$ equals $6\, \text{cm}$ and a height $h$ equals $8\, \text{cm}$ is ().\\
\textbf{Choices:} (A) 94.2 (B) 150.8 (C) 188.5 (D) 251.3 \\
 \textbf{Model response:} Given $h = 8\, \text{cm}$ and $r = 6\, \text{cm}$, the length of the generatrix of the cone can be set as $l$. Using the Pythagorean theorem, we have $l = \sqrt{6^2 + 8^2} = 10\, \text{cm}$. The lateral area of the unfolded cone is $S_{\text{side}} = 0.5 \times 2 \times 6\pi \times 10 = 60\pi\, \text{cm}^2$. Therefore, the lateral area of the cone is $60\pi\, \text{cm}^2$. Therefore, the answer is C.
\\
 \textbf{Extracted answer:} \textbf{C}
 \end{minipage}
 \\
 \midrule
 Example 3 & 
 \begin{minipage}[s][2.6cm]{1.5\columnwidth}
 \textbf{Question:} In triangle ABC, F is the midpoint of BC and point E is on the AC side. AC = 10. What is the length of AE?\\
 \textbf{Choices:} (A) 3.0 (B) 4.0 (C) 5.0 (D) 4.5\\
 \textbf{Model response:} Since F is the midpoint of BC, EF is parallel to AB, so EF is the median of triangle ABC. Therefore, point E is the midpoint of AC. Therefore, AE = 0.5 × AC. Since AC = 10, AE = 5. Therefore, the answer is C.\\
 \textbf{Extracted answer:} \textbf{C}
 \end{minipage}
 \\
 \bottomrule
\end{tabular} 
\caption{Task-specific instructions for extracting the answer. The table shows three examples with answers that can be extracted using the prompt.}

\label{tab:promt_answer_extraction}
\end{table*}

\begin{table*}[th]
\centering
\small
\begin{tabular}{ll}
\toprule
\textbf{Regular expressions} & \textbf{Demonstration Examples} \\
\midrule
\verb|value of (\w+) is\s*([\d.]+)| & The value of x is 3.5. \\
\verb|correct answer is\s*(.+).| & correct answer is C." \\
\verb|answer is\s*([\d.]+)| & answer is 17.1." \\
\verb|answer should be\s*(.+) degrees| & Therefore, the answer should be choice D." \\
\verb|answer to (.+) is (.+) degrees| & The answer to the angle ABC is $60^\circ$ \\
\verb|answer to the problem is\s*(.+)| & The correct answer to problem is $y = x^2 + 2x + 3$." \\
\verb|The closest (.+) is (.+).| & So we got the area is 13.1. The closest answer is D." \\
\verb|the (.+) is equal to (.+).| & The degree measure of angle ABC is 35 degrees. \\
\verb|(.+) is approximately (.+) units| & So, the length of the line segment is approximately 10 units." \\
\bottomrule
\end{tabular}
\caption{Regular expressions used for extracting the answers that first step extraction cannot handle. The "Demonstration Examples" columns display corresponding examples that the regular expressions can match.}
\label{tab: human_ext_rules}
\end{table*}

\section{Statistic Analysis}
\label{sec:appendix-statistical-analysis}

\begin{table}[tbhp!]
\centering
\begin{small}
\begin{tabular}{@{}lc@{}}
\toprule
\textbf{Total Numbers}         & \multicolumn{1}{l}{} \\
- GeoEval-2000                 & 2,000                \\
- GeoEval-backward             & 750                  \\
- GeoEval-aug                  & 2,000                \\
- GeoEval-hard                 & 300                  \\ \midrule
\textbf{Input Types}           &                      \\
- text + description           & 1,120                \\
- text + diagram               & 1,120                \\
- text + description + diagram & 1,166                \\ \midrule
\textbf{Answer Types}          &                      \\
- number                       & 5,050                \\
- expression                   & 232                  \\
- coordinate                   & 68                   \\ \midrule
\textbf{Problem Types}         &                      \\
- flat geometry                & 5,050                \\
- solid geometry               & 272                  \\
- analytic geometry            & 28                   \\ \midrule
\textbf{Others}                &                      \\
- average problem length      & 28                   \\
- average description length   & 34                   \\
- geometry shapes              & 12                   \\ \bottomrule
\end{tabular}
\end{small}
\caption{Statistics of GeoEval benchmark.}
\label{tab:statistics}
\end{table}

Table~\ref{tab:statistics} presents a statistical breakdown of the GeoEval benchmark. This benchmark encompasses a total of 5,050 geometry math problems, categorized into four subsets: GeoEval-2000 (2,000 problems), GeoEval-backward (750 problems), GeoEval-aug (2,000 problems), and GeoEval-hard (300 problems). Besides the problem text, each problem in the dataset includes at least one of the following: a geometric diagram, a description of the diagram, or both. The majority of the correct answers are numerical values, with a minority comprising expressions, coordinates, or option letters, primarily in the GeoEval-hard subset.

\section{Source Datasets}
\label{sec:appendix-source-datasets}

\begin{figure}[tbhp!]
\centering
\includegraphics[width=0.45\textwidth]{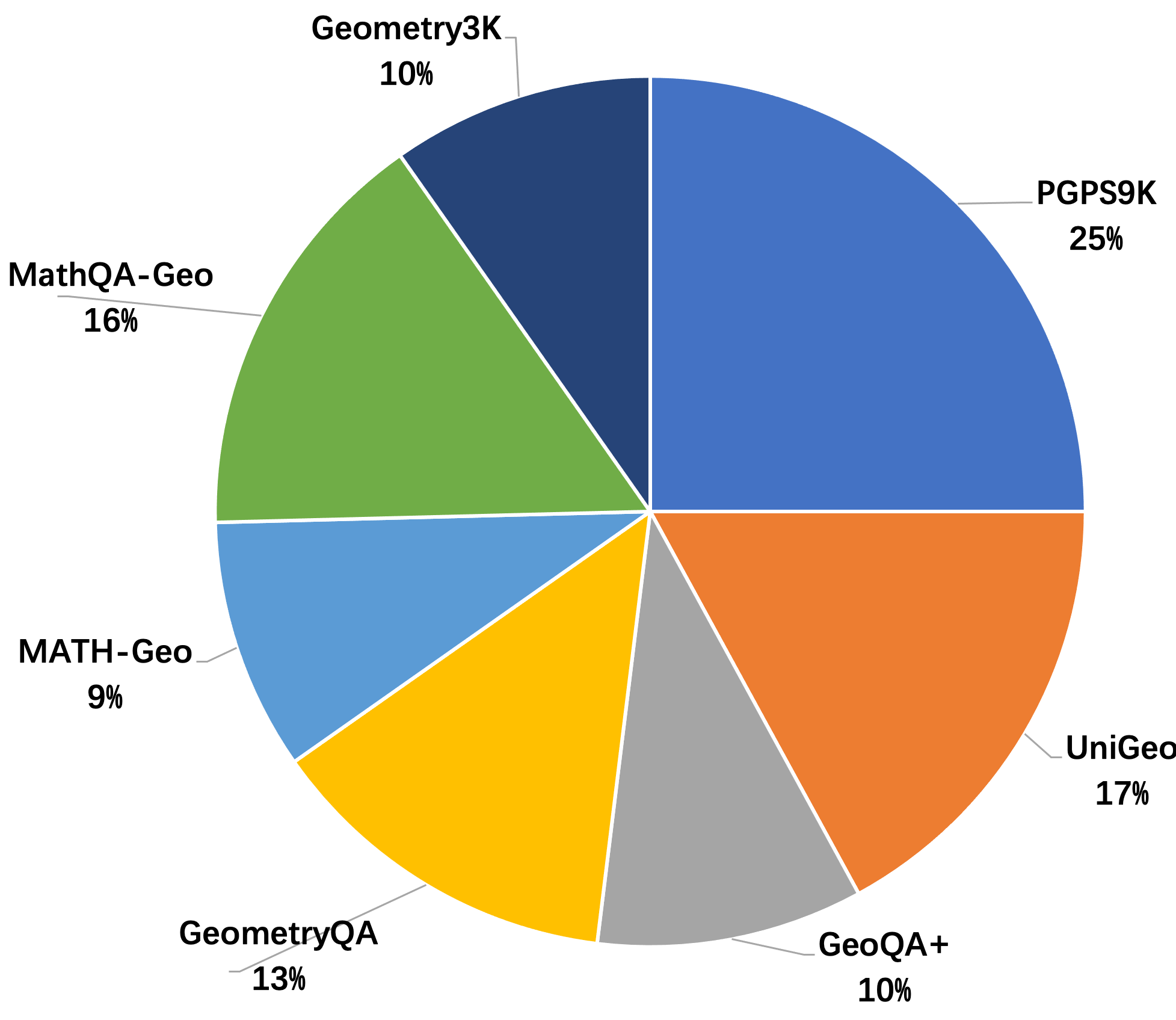}
\caption{Distributions of source datasets.}
\label{fig:distribution-of-source-datasets}
\end{figure}

Table~\ref{tab:source-datasets-details} provides details on the source datasets that contribute to the GeoEval-2000 subset, including information on their content and characteristics. Meanwhile, Figure~\ref{fig:distribution-of-source-datasets} visualizes the proportional contributions of these source datasets to the GeoEval-2000 subset, showcasing the variety and scope of the geometry problems collected from each source.

\begin{table}[tbhp!]
\centering
\begin{scriptsize}
\begin{tabular}{@{}lccc@{}}
\toprule
Source Dataset & Diagram & Diagram Descriptions & Quantity \\ \midrule
Geometry3K     & \cmark     & \cmark                  & 3001     \\ \midrule
PGPS9K         & \cmark     & \cmark                  & 9022     \\ \midrule
UniGeo         & \cmark     & \xmark                   & 4998 $\dagger$     \\ \midrule
GeoQA+         & \cmark     & \xmark                   & 2518     \\ \midrule
GeometryQA     & \xmark      & \xmark                   & 1398     \\ \midrule
MATH           & \xmark      & \xmark                   & 1349 $\ddagger$  \\ \midrule
MathQA         & \xmark      & \xmark                   & 2625 $\ddagger$     \\ \bottomrule
\end{tabular}
\end{scriptsize}
\caption{The information of source datasets for GeoEval-2000 dataset. The "$\dagger$" symbol indicates that proving problems from the UniGeo dataset have been excluded. The "$\ddagger$" sign specifies that the count only pertains to geometry problems within the dataset, focusing on problems directly relevant to the GeoEval-2000's scope.}
\label{tab:source-datasets-details}
\end{table}

\section{Distributions of Different Geometric Shapes}
\label{sec:appendix-distributions-of-different-geometric-shapes}

\begin{figure}[tbhp!]
\centering
\includegraphics[width=0.45\textwidth]{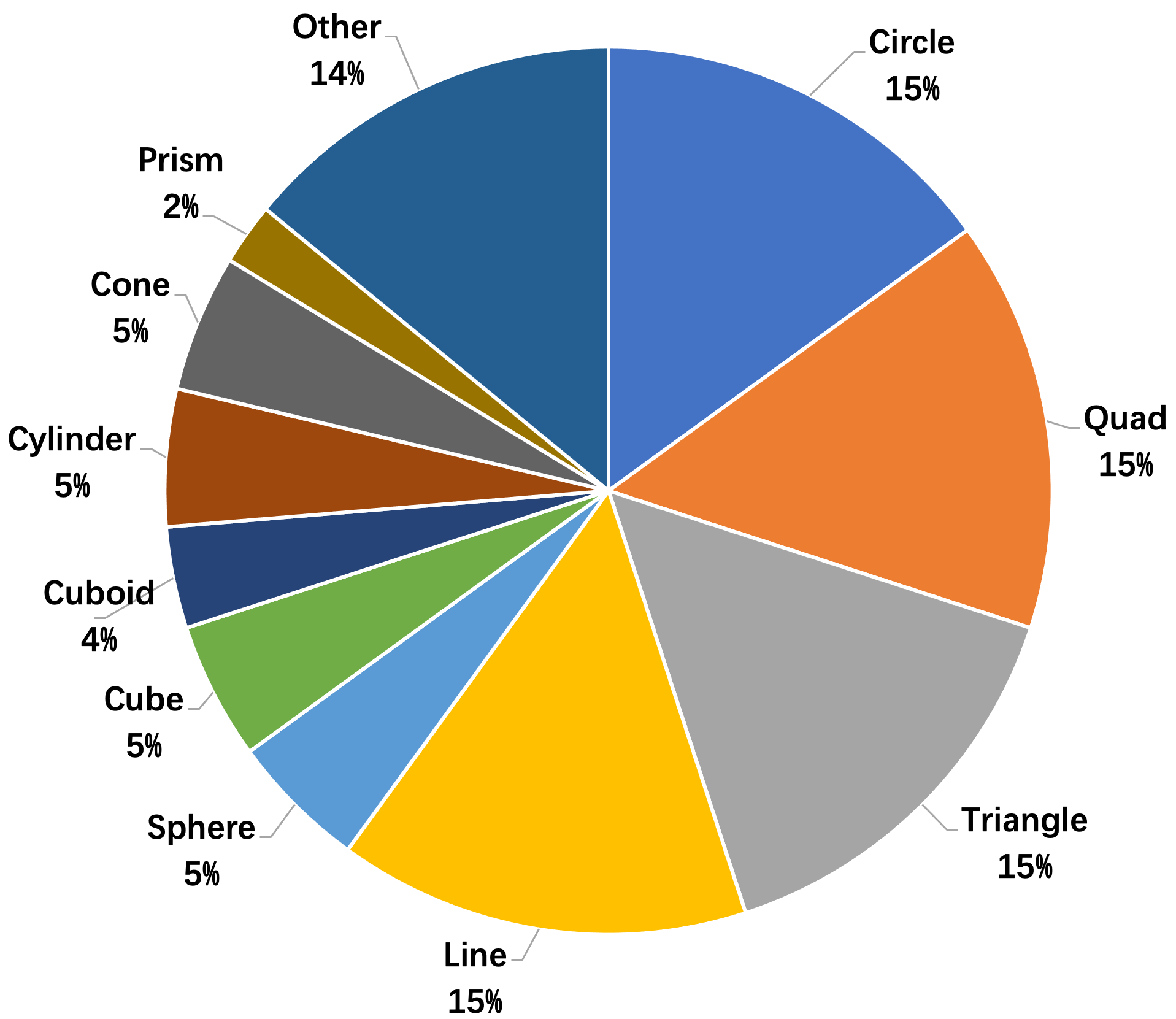}
\caption{Distributions of different geometric shapes.}
\label{fig:distribution-geometric-shapes}
\end{figure}

Figure~\ref{fig:distribution-geometric-shapes} illustrates the varied distribution of geometric shapes within the GeoEval-2000 subset, highlighting the diversity of geometry concepts represented in this collection.

\section{Backward Question Example}
\label{sec:appendix-backward-question-example}

\begin{figure}[tbhp!]
\centering
\includegraphics[width=0.48\textwidth]{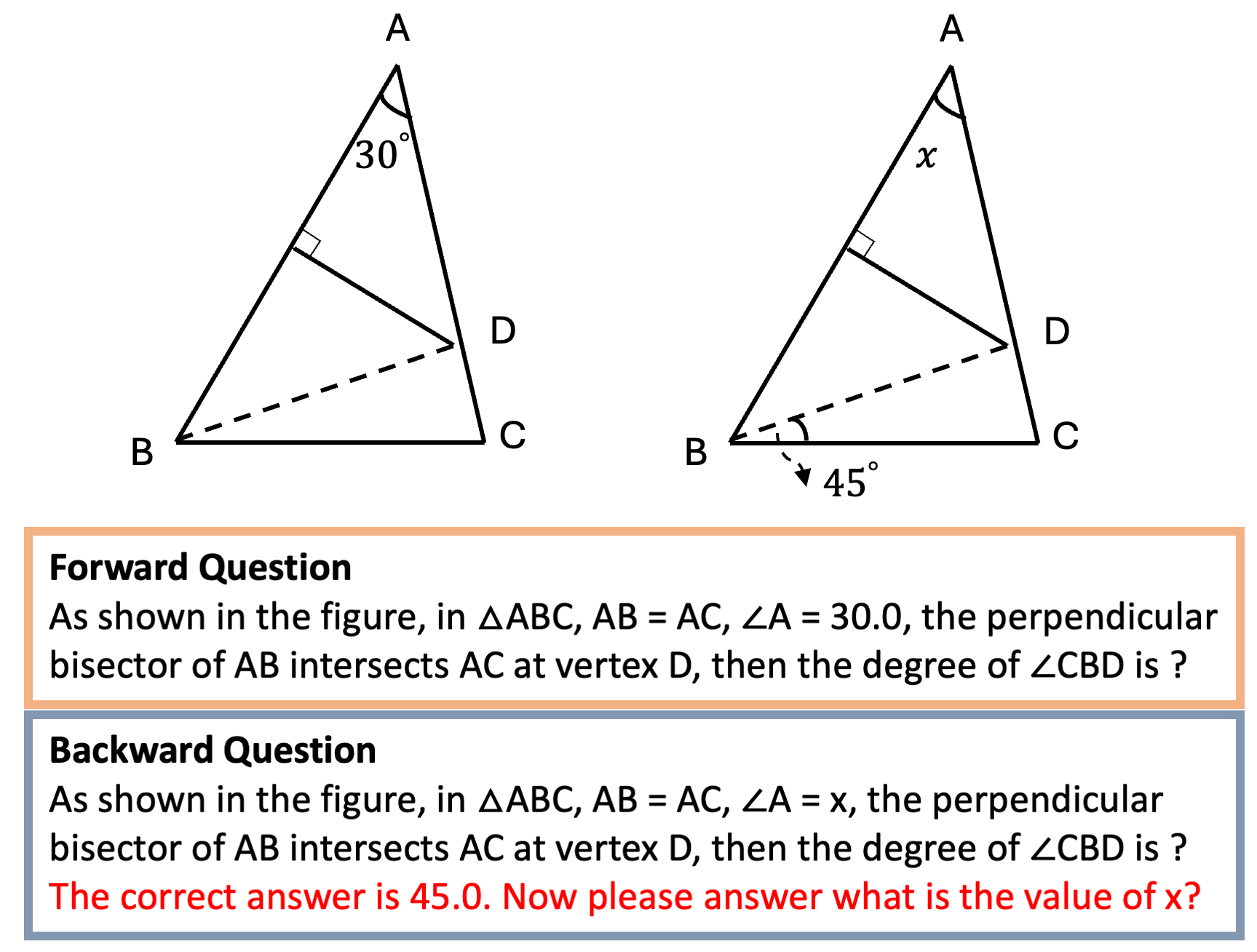}
\caption{Example for the backward question. The left and right figures are diagrams for the forward question and backward question, respectively.}
\label{fig:backward-example}
\end{figure}

Figure~\ref{fig:backward-example} is an example from the GeoEval-backward subset.

% \begin{figure}[tbhp!]
% \centering
% \includegraphics[width=0.45\textwidth]{images/backward.png}
% \caption{Example for the backward question.}
% \label{fig:backward-example}
% \end{figure}

\section{Comparison between GeoEval-hard subset and other public datasets}
\label{sec:appendix-comparison-between-geohard-subset-and-other-public-datasets}

\begin{table*}[tbhp!]
\centering 
\begin{small}
\setlength{\tabcolsep}{1pt}

\resizebox{0.95\linewidth}{!}{
\begin{tabular}{l|ccccc}
	\toprule
        \multirow{2}{*}{\textbf{Dataset}} & \multicolumn{2}{c}{Solid Geometry} & \multicolumn{3}{c}{Analytic Geometry} \\
        \cmidrule(lr){2-3} \cmidrule(lr){4-6}
        & \#solid geometry shapes & \#question type   & \#geometry curve knowledge & \#question types & \#grade  \\
	\midrule
        UniGeo \cite{unigeo} & \xmark & calculate/prove  & \xmark & -- & 6-12 \\
        GeoQA \cite{geoqa+} & \xmark &  calculate  & \xmark & -- & 6-12 \\
        Geometry3K \cite{geometry3k} & \xmark & calculate & \xmark & --  & 6-12 \\
	PGPS9K \cite{pgps9k} & \xmark & calculate/judge  & \xmark & -- & 6-12 \\
	MathVista(Geometry Part) \cite{MathVista} & \xmark & calculate/judge & \xmark & -- & -- \\
        MathVista(FunctionQA Part) \cite{MathVista} & \xmark & calculate/judge & \cmark & judge & -- \\
        \midrule
	\textbf{GeoEval-hard} & \cmark & judge/calculate/reason & \cmark & judge/calculate/reason & 9-12  \\
	\bottomrule
\end{tabular}
}
\end{small}
\vspace{0mm}
\caption{Comparison between GeoEval-hard with other public datasets.}
\vspace{0mm}
\label{tab:sa-dataset}
\end{table*}

To thoroughly assess the models' abilities in grasping concepts of solid and analytic geometry, the GeoEval-hard subset was created to include a diverse range of visual elements, such as three-dimensional views, across a spectrum of topics in solid geometry. The distinctions between the GeoEval-hard subset and other publicly available datasets are detailed in Table~\ref{tab:sa-dataset}, demonstrating the unique coverage and complexity of the GeoEval-hard subset in comparison.

\section{Inspection of GeoEval-hard subset}
\label{sec:appendix-data-correctness-check-for-geoeval-hard}

To ensure the GeoEval-hard dataset's high quality and accuracy, and prevent the LLMs and MMs from recalling or inferring solutions to problems with similar structures or templates as those seen during training, we made several modifications to the original problems sourced from the GeoEval-hard-raw subset, where these modifications were made to prevent direct recall or plagiarism from the training data, while still preserving the underlying mathematical reasoning required. Specifically, we have adopted these steps:

\begin{enumerate}
    \item We changed the variable names and numerical values used in the problem statements.
    \item We swapped the original question with the provided conditions.
    \item We used the original answer as a new condition in the reformulated problem statement.
\end{enumerate}

Finally, we thoroughly analyze the revised reformulated problem to ensure it can be solved. Specifically, we form a team of six reviewers, each holding at least a Master's degree, to scrutinize every question. This evaluation process is structured in three phases: individual review, swap review, and candidate review. The primary focus lies on two key standards: the completeness and relevance of the geometric diagrams, and the reasonableness of the answers provided.

In the first phase, "individual review", each reviewer is randomly assigned 50 geometry math problems from the GeoEval-hard dataset. Their task is to assess the geometry math problems based on the standards, marking any that fail to meet these standards. During the "swap review" phase, these sets of 50 geometry math problems are exchanged among reviewers for a second evaluation. To ensure unbiased assessment, we hide the results of the initial review. Here, reviewers again highlight geometry math problems not conforming to the standards. The final phase, "candidate review", involves selecting geometry math problems for the dataset based on the outcomes of the first two phases. Geometry math problems unmarked in both phases are retained, those marked in both are discarded, and those highlighted in only one phase undergo further examination by the entire review team, with the majority decision determining their inclusion.

\section{Examples from GeoEval Representing Five Features}
\label{sec:appendix-examples-from-geoeval-representing-five-features}

\subsection{Comprehensive Variety}

\begin{figure*}[tbhp!]
\centering
\includegraphics[width=1.0\textwidth]{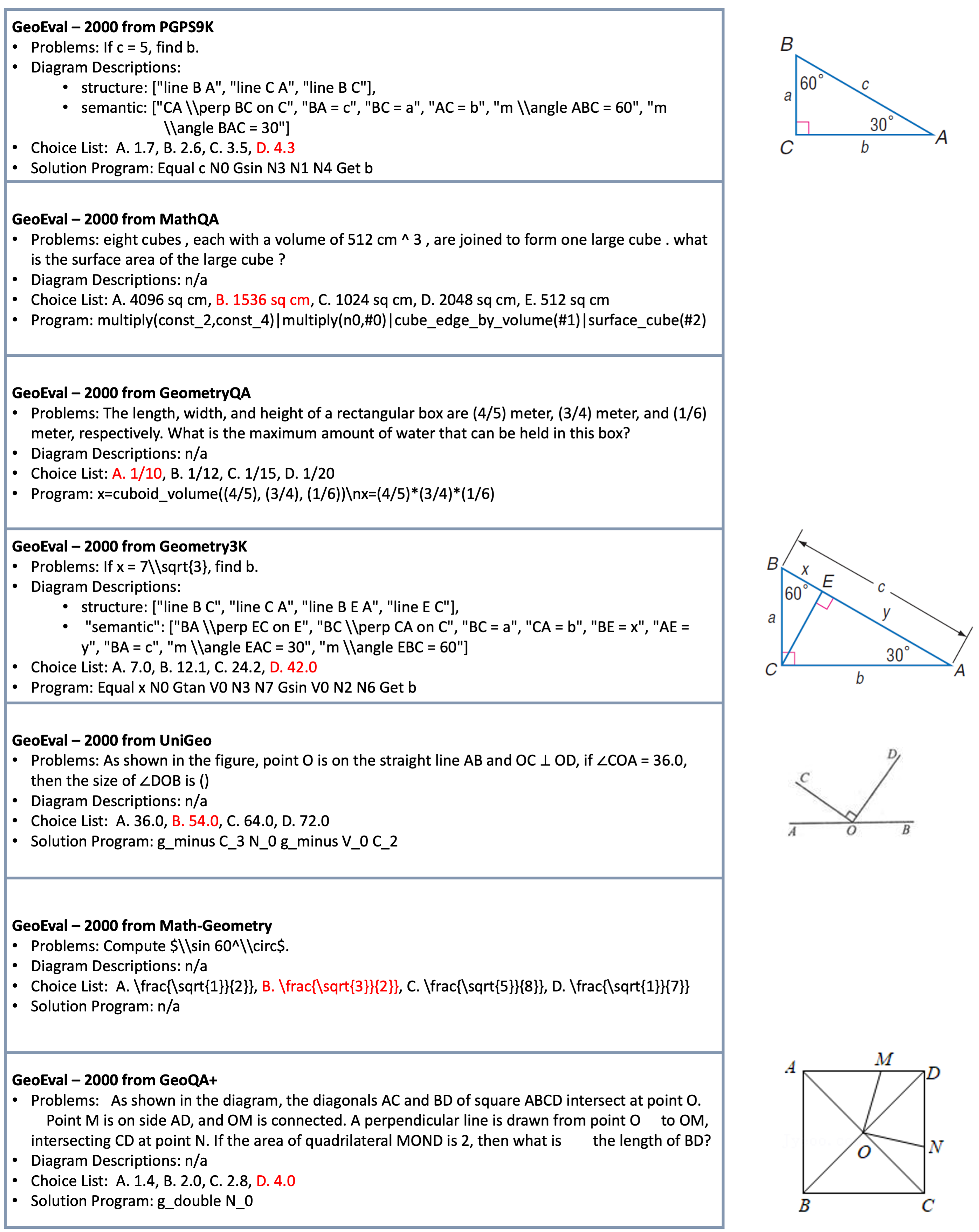}
\caption{Examples from GeoEval-2000 dataset. The golden answer choice is highlighted in red color.}
\label{fig:feature_examples_1}
\end{figure*}

Figure~\ref{fig:feature_examples_1} presents sample data from the GeoEval-2000 subset, illustrating its diversity in terms of data sources.

\subsection{Varied Problems}

\begin{figure*}[tbhp!]
\centering
\includegraphics[width=1.0\textwidth]{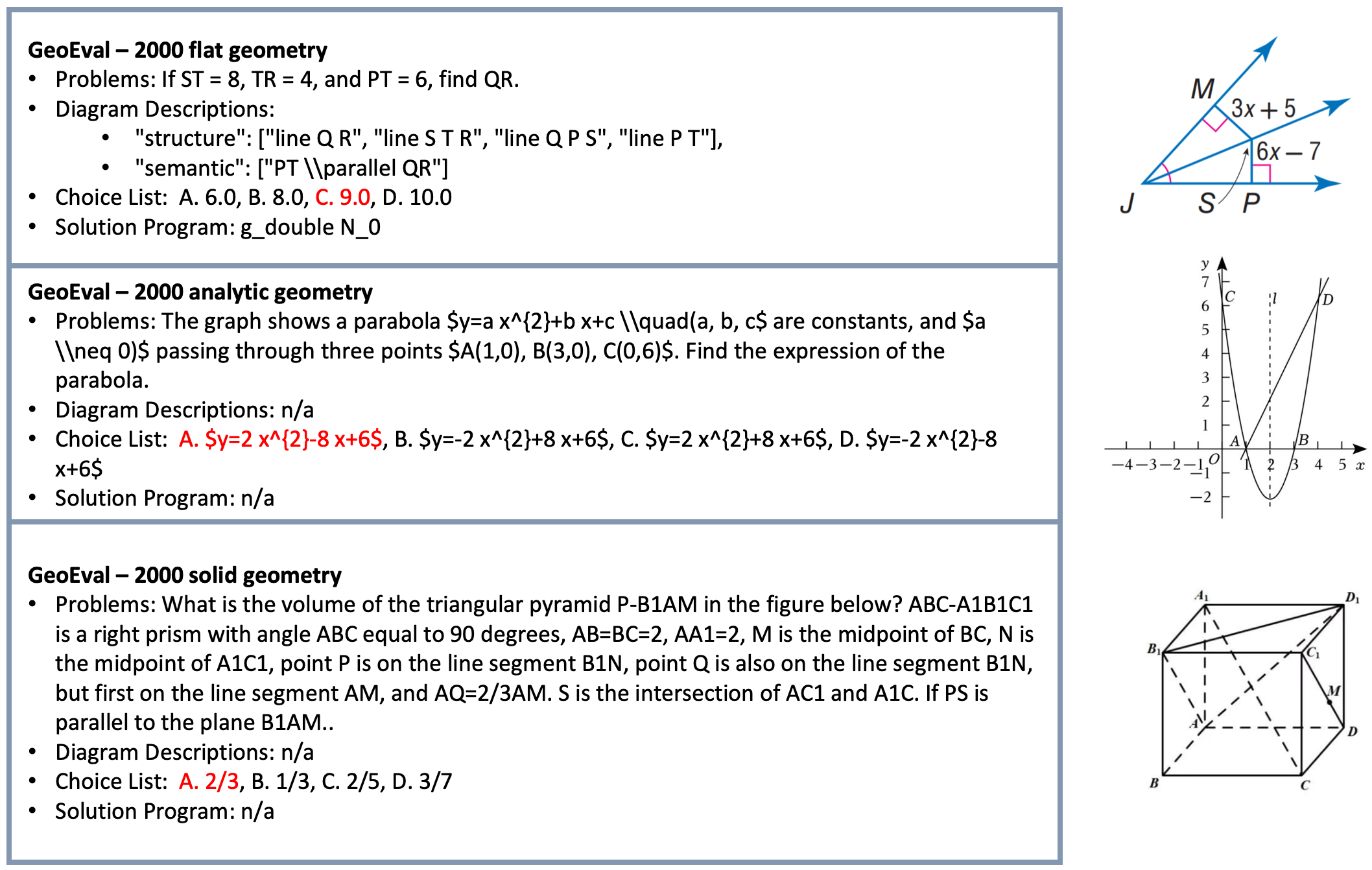}
\caption{Examples of the flat geometry problem, the analytic geometry problem, and the solid geometry problem in GeoEval benchmark.}
\label{fig:varied_problems}
\end{figure*}

Figure~\ref{fig:varied_problems} displays examples of three distinct problem types in the GeoEval benchmark: flat geometry, analytic geometry, and solid geometry.

\subsection{Dual Inputs}

Figure~\ref{fig:feature_examples_1} shows that the GeoEval benchmark comprises geometry math problems that contain both diagrams and textual descriptions, as well as problems that include textual descriptions alone.

\subsection{Diverse Challenges}

\begin{figure*}[tbhp!]
\centering
\includegraphics[width=1.0\textwidth]{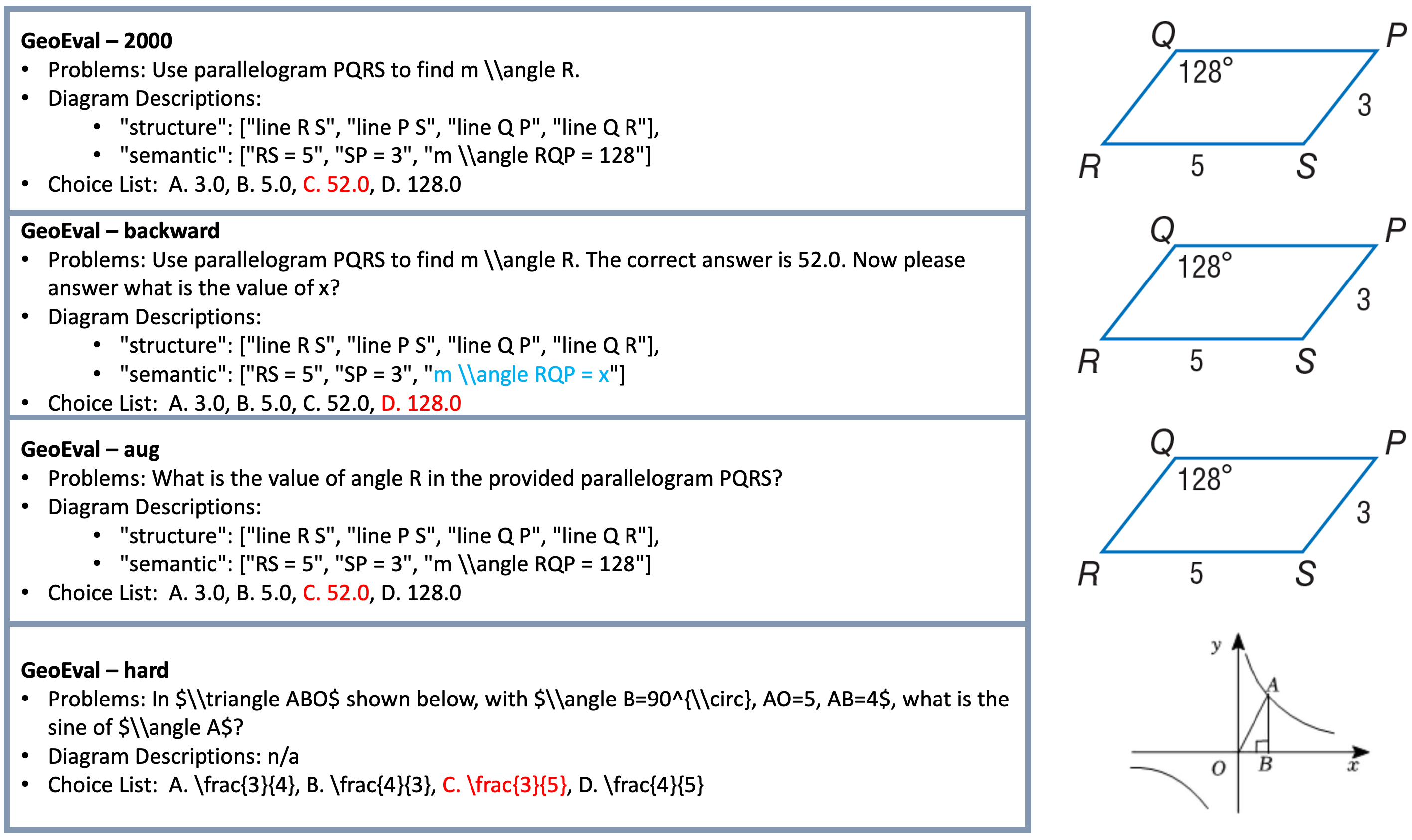}
\caption{Examples of GeoEval-2000, GeoEval-backward, GeoEval-aug, GeoEval-hard subsets.}
\label{fig:ood}
\end{figure*}

Figure~\ref{fig:ood} showcases examples from the GeoEval-2000, GeoEval-backward, GeoEval-aug, and GeoEval-hard subsets, illustrating the diverse challenges within the GeoEval benchmark.

\subsection{Complexity Ratings}

\begin{figure*}[tbhp!]
\centering
\includegraphics[width=1.0\textwidth]{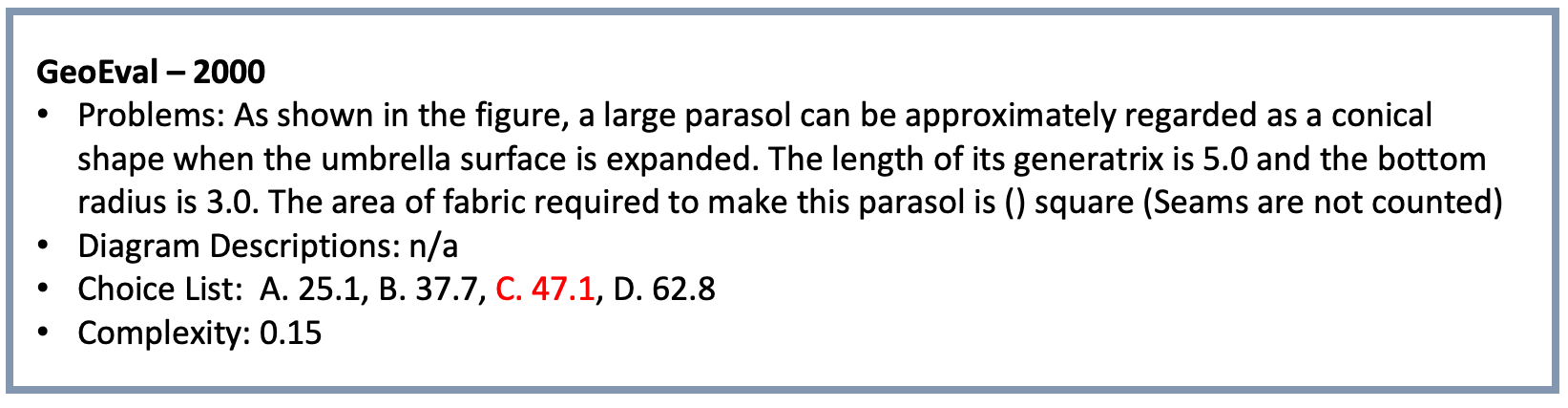}
\caption{Example for a problem annotated with complexity in the GeoEval benchmark.}
\label{fig:complexity}
\end{figure*}

Every problem in the GeoEval benchmark is annotated with a complexity rating, indicating the level of skill necessary to solve it, as shown in Figure~\ref{fig:complexity}.

\section{Algorithm for Classifying Geometry Math Problems Complexity}
\label{sec:appendix-complexity}

Algorithm~\ref{alg:complexity} details our methodology for classifying each geometry math problem into distinct levels of complexity.

\section{Evaluation Details}
\label{sec:appendix-evaluation-details}

\subsection{Model Hyper-parameters}

% \usepackage{booktabs}
\begin{table*}[tbhp!]
\centering
\begin{scriptsize}
\begin{tabular}{@{}lll@{}}
\toprule
\multicolumn{1}{c}{Model Name} & \multicolumn{1}{c}{Generation Parameters}                & \multicolumn{1}{c}{Comments}     \\ \midrule
CodeGen2-16B                   & do\_sample=True, top\_k=0.5, top\_p=0.5, max\_tokens=512 & model=""Salesforce/codegen2-16B" \\ \midrule
WizardMath-7B-V1.1             & temperature=0.0, top\_p=1, max\_tokens=1024              & vLLM package                     \\ \midrule
WizardMath-70B                 & temperature=0.0, top\_p=1, max\_tokens=1024              & vLLM package                     \\ \midrule
GPT-3.5                        & temperature=0.7, max\_tokens=512                         & version="gpt-3.5-turbo-0125"     \\ \midrule
GPT-4                          & temperature=0.7, max\_tokens=512                         & version="gpt-4-1106-preview"     \\ \midrule
llava-7B-V1.5                  & temperature=0.0, max\_new\_tokens=512                    & llava package                    \\ \midrule
Qwen-VL                        & temperature=0.0, max\_new\_tokens=512                    & model="Qwen/Qwen-VL"             \\ \midrule
mPLUG-Owl2                     & do\_sample=True, top\_p=0.7, max\_tokens=512             & model="mPLUG-Owl2"               \\ \midrule
InstructBLIP & do\_sample=False, num\_beams=5, max\_tokens=512, top\_p=0.9, temperature=1.0 & model="Salesforce/instructblip-vicuna-7b" \\ \midrule
GPT-4V                         & temperature=0.0, max\_tokens=512                         & version="gpt-4-vision-preview"   \\ \bottomrule
\end{tabular}
\caption{The hyper-parameters for the models used in the evaluation are detailed. When the "comments" section includes the format \textit{model = ""}, it signifies that the model was loaded from the transformer package. The vLLM package indicates that models are implemented by the vLLM package, where more details can be found in \url{https://github.com/vllm-project/vllm}. For models other than OpenAI's GPT, custom codes were utilized for evaluation unless specified otherwise in the comments.}
\label{tab:model-hyperparameters}
\end{scriptsize}
\end{table*}

Table~\ref{tab:model-hyperparameters} presents the complete list of hyper-parameters applied to the models throughout the evaluation phase.

\subsection{Instruction Prompt Used for Evaluating Models}

% \usepackage{booktabs}
\begin{table*}[]
\centering
\begin{scriptsize}
\begin{tabular}{@{}lll@{}}
\toprule
 &
  Template &
  Example \\ \midrule
Merge &
  \begin{tabular}[c]{@{}l@{}}Here are the basic description of the diagram: \$\{diagram descriptions\},\\ \$\{problems texts\},\\ The Choices are: \$\{choice list\}\end{tabular} &
  \begin{tabular}[c]{@{}l@{}}Please solve this math problem:\\ Here are the basic description of the diagram:  \\ line B A, line C A, line B C\textbackslash{}nCA \textbackslash{}\textbackslash{}perp BC on C, \\ BA = c, BC = a, AC = b, m \textbackslash{}\textbackslash{}angle ABC = 60, \\ m \textbackslash{}\textbackslash{}angle BAC = 30\textbackslash{}nIf c = 5, \\ find b.\\ The Choices are:  {[}1.7, 2.6, 3.5, 4.3{]}\end{tabular} \\ \midrule 
Instruction &
  \begin{tabular}[c]{@{}l@{}}Please solve this math problem:\\ \$\{Merge\}\\ \#\#\# Problem-solving Bot:\end{tabular} &
  \begin{tabular}[c]{@{}l@{}}Please solve this math problem:\\ Here are the basic description of the diagram:  \\ line B A, line C A, line B C\textbackslash{}nCA \textbackslash{}\textbackslash{}perp BC on C, \\ BA = c, BC = a, AC = b, m \textbackslash{}\textbackslash{}angle ABC = 60, \\ m \textbackslash{}\textbackslash{}angle BAC = 30\textbackslash{}nIf c = 5, \\ find b.\\ The Choices are:  {[}1.7, 2.6, 3.5, 4.3{]}\\ \#\#\# Problem-solving Bot:\end{tabular} \\ \bottomrule
\end{tabular}
\end{scriptsize}
\caption{Templates and examples provided illustrate the process of merging and instruction creation. The placeholder "\$\{Merge\}" represents the combined texts of "diagram descriptions," "problems texts", and "choice list". In cases where "diagram descriptions" are absent, the phrase "Here are the basic description of the diagram:" is omitted.}
\label{tab:instruction-prompt}
\end{table*}

Before employing instruction prompts to steer model responses, we combine the problem texts, diagram descriptions, and choice lists from an example, as depicted in the "Merge" row of Table~\ref{tab:instruction-prompt}. Following this combination, as illustrated in the "Instruction" row of Table~\ref{tab:instruction-prompt}, we incorporate instruction prompts into the merged texts and then forward these to the models to generate responses.

% \section{Prompt \& Heuristic rules For Answer Extraction}
% \label{sec:appendix-answer-extraction}

% We detail the prompts utilized for extraction using GPT-4, which include an extraction instruction alongside various sample prompts. The extraction instruction and the constructed samples are presented in Table \ref{tab:promt_answer_extraction} and Table \ref{tab: human_ext_rules}, illustrating the methodology behind the extraction process. \textcolor{blue}{To clarify, our extraction pipeline involves two steps: Firstly, using a prompt to extract the answer, where Table 9 illustrates with three examples, showcasing the prompt and extracted answer formats. Secondly, employing regular expressions to extract any remaining answers that couldn't be obtained from the first step. As shown in Table 10, it offers an example for each regular expression, aiding in understanding the regex patterns.}

\section{Reason for Removing InstructBLIP from the Comparison}
\label{sec:appendix-reason-for-removing-instructblip}

\begin{figure*}[tbhp!]
\centering
\includegraphics[width=1.0\textwidth]{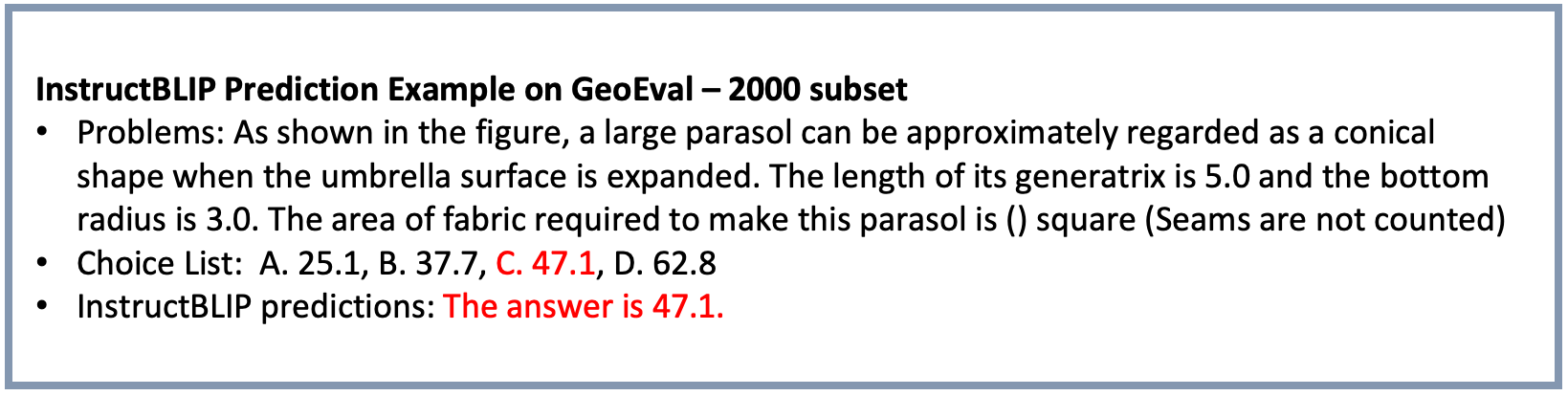}
\caption{One example of InstructBLIP prediction on GeoEval-2000 subset.}
\label{fig:instructblip}
\end{figure*}

\begin{table*}[tbhp!]
\centering
\begin{scriptsize}
\begin{tabular}{@{}lcccccc@{}}
\toprule
 &
  \multicolumn{6}{c}{GeoEval-2000 Data Sources} \\
\multicolumn{1}{l|}{Models} &
  \multicolumn{1}{c|}{MATH (Geometry) (\%)} &
  \multicolumn{1}{c|}{GeometryQA (\%)} &
  \multicolumn{1}{c|}{GeoQA+ (\%)} &
  \multicolumn{1}{c|}{PGPS9K (\%)} &
  \multicolumn{1}{c|}{UniGeo (\%)} &
  MathQA (Geometry) (\%) \\ \midrule
\multicolumn{1}{l|}{CodeGen2-16B} &
  \multicolumn{1}{c|}{0.36} &
  \multicolumn{1}{c|}{0.35} &
  \multicolumn{1}{c|}{0.44} &
  \multicolumn{1}{c|}{0.18} &
  \multicolumn{1}{c|}{0.41} &
  0.25 \\
\multicolumn{1}{l|}{GPT-3.5} &
  \multicolumn{1}{c|}{0.35} &
  \multicolumn{1}{c|}{0.31} &
  \multicolumn{1}{c|}{0.19} &
  \multicolumn{1}{c|}{0.27} &
  \multicolumn{1}{c|}{0.23} &
  0.26 \\
\multicolumn{1}{l|}{GPT-4} &
  \multicolumn{1}{c|}{0.58} &
  \multicolumn{1}{c|}{0.74} &
  \multicolumn{1}{c|}{0.27} &
  \multicolumn{1}{c|}{0.28} &
  \multicolumn{1}{c|}{0.27} &
  0.44 \\ \midrule
\multicolumn{1}{l|}{WizardMath-7B-V1.1} &
  \multicolumn{1}{c|}{0.58} &
  \multicolumn{1}{c|}{0.53} &
  \multicolumn{1}{c|}{0.59} &
  \multicolumn{1}{c|}{0.55} &
  \multicolumn{1}{c|}{0.54} &
  0.35 \\
\multicolumn{1}{l|}{WizardMath-70B} &
  \multicolumn{1}{c|}{0.54} &
  \multicolumn{1}{c|}{0.58} &
  \multicolumn{1}{c|}{0.62} &
  \multicolumn{1}{c|}{0.54} &
  \multicolumn{1}{c|}{0.57} &
  0.35 \\ \midrule
\multicolumn{1}{l|}{llava-7B-V1.5} &
  \multicolumn{1}{c|}{0.26} &
  \multicolumn{1}{c|}{0.4} &
  \multicolumn{1}{c|}{0.12} &
  \multicolumn{1}{c|}{0.15} &
  \multicolumn{1}{c|}{0.12} &
  0.19 \\
\multicolumn{1}{l|}{Qwen-VL} &
  \multicolumn{1}{c|}{0.29} &
  \multicolumn{1}{c|}{0.46} &
  \multicolumn{1}{c|}{0.27} &
  \multicolumn{1}{c|}{0.22} &
  \multicolumn{1}{c|}{0.32} &
  0.24 \\
\multicolumn{1}{l|}{mPLUG-Owl2} &
  \multicolumn{1}{c|}{0.27} &
  \multicolumn{1}{c|}{n/a} &
  \multicolumn{1}{c|}{0.29} &
  \multicolumn{1}{c|}{0.46} &
  \multicolumn{1}{c|}{0.27} &
  0.0 \\
\multicolumn{1}{l|}{InstructBLIP} &
  \multicolumn{1}{c|}{0.0} &
  \multicolumn{1}{c|}{n/a} &
  \multicolumn{1}{c|}{0.59} &
  \multicolumn{1}{c|}{0.48} &
  \multicolumn{1}{c|}{0.57} &
  0.0 \\
\multicolumn{1}{l|}{GPT-4V} &
  \multicolumn{1}{c|}{0.45} &
  \multicolumn{1}{c|}{0.61} &
  \multicolumn{1}{c|}{0.34} &
  \multicolumn{1}{c|}{0.38} &
  \multicolumn{1}{c|}{0.45} &
  0.38 \\ \bottomrule
\end{tabular}
\end{scriptsize}
\caption{The accuracy scores achieved by models on different sources datasets constituting the GeoEval-2000 subset.}
\label{tab:different-sources}
\end{table*}

As shown in Figure~\ref{fig:instructblip}, InstructBLIP's responses on the GeoEval-2000 subset are typically scalar, lacking any intermediate reasoning steps. This suggests that InstructBLIP may have been exposed to GeoEval-2000 questions during its pre-train phase, leading to the memorization of answers. This is supported by the observed performance decline from GeoEval-2000 to GeoEval-aug, which falls from 52.18\% to 35.00\%. Additionally, InstructBLIP tends to directly generate option letters (e.g., "A") for the GeoEval-hard subset without any reasoning process, resulting in an improbably high accuracy rate of 70.30\% for this subset. Consequently, in our subsequent analysis and discussions, we have chosen to exclude the InstructBLIP model.

\section{Models Performances across Different Data Sources}
\label{sec:appendix-data-sources}

Table~\ref{tab:different-sources} shows model performances on the GeoEval-2000 subset according to the different original datasets. We can observe that WizardMath models still achieve the best accuracy scores on almost all datasets.

\SetKwComment{Comment}{/* }{ */}
\begin{algorithm*}[tbhp!]
\caption{Algorithm for classifying geometry math problems complexity}
\label{alg:complexity}
\KwIn{$\text{All Problem Texts } T, \text{All Diagram Descriptions } D, \text{All Golden Solution Programs } S$}
\KwOut{Complexity for each problem}
$len_{T,D} = 0$\;
$len_{S} = 0$\;
\For{$t$ \textbf{ in } $T$, $d$ \textbf{ in } $D$, $s$  \textbf{ in } $S$}{
    $len_{T,D} += len_{t} + len_{d} $ \Comment*[r]{sum up the length of problem texts and the length of diagram descriptions.}
    $len_{S} += len_{s}$ \Comment*[r]{sum up the length of golden solution programs.}
}

\For{$t$ \textbf{ in } $T$, $d$ \textbf{ in } $D$, $s$  \textbf{ in } $S$}{
    $C_{t,d,s} \leftarrow \alpha \times \frac{len_{t}+len_{d} - \text{min}(len_{T,D})}{\text{max}(len_{T,D}) - \text{min}(len_{T,D})} + (1 - \alpha) \times \frac{len_{s} - \text{min}(len_{S})}{\text{max}(len_{S}) - \text{min}(len_{S})}$\;

    \uIf{$0.0 \leq C_{t,d,s} \leq 0.2$}{
        $\text{Complexity} \leftarrow \text{Easy}$\Comment*[r]{classify the problem as Easy problem.}
    }
    \uElseIf{$0.2 < C_{t,d,s} \leq 0.6$}{
        $\text{Complexity} \leftarrow \text{Middle}$\Comment*[r]{classify the problem as Middle problem.}
    }
    \ElseIf{$0.6 < C_{t,d,s} \leq 1.0$}{
        $\text{Complexity} \leftarrow \text{Hard}$\Comment*[r]{classify the problem as Hard problem.}
    }
}
\end{algorithm*}

\end{document}